\documentclass[3p,times,authoryear]{elsarticle}
\usepackage{ecrc}
\usepackage{amsmath,amsfonts}
\usepackage{algorithmic}
\usepackage{algorithm}
\usepackage{array}
\usepackage{amssymb}
\usepackage[caption=false,font=normalsize,labelfont=sf,textfont=sf]{subfig}
\usepackage{textcomp}
\usepackage{stfloats}
\usepackage{url}
\usepackage{verbatim}
\usepackage{graphicx}
\usepackage[table]{xcolor}
\usepackage{diagbox}
\usepackage{multirow}
\usepackage{color}
\usepackage{booktabs}
\usepackage{pifont}
\usepackage[backref,pagebackref=false]{hyperref}
\hypersetup{
hidelinks,
colorlinks=true,
linkcolor=black,
citecolor=black
}

\volume{00}

\firstpage{1}

\journalname{Neural Networks}

\runauth{}

\jid{procs}

\jnltitlelogo{Procedia Computer Science}

\CopyrightLine{2011}{Published by Elsevier Ltd.}

\usepackage{amssymb}

\usepackage[figuresright]{rotating}

\begin{document}
\begin{frontmatter}



\dochead{}

\title{GSB: Group Superposition Binarization for Vision Transformer with Limited Training Samples}

\author[1]{Tian Gao}
\ead{gaotian970228@njust.edu.cn}
\address[1]{School of Computer Science and Engineering, Nanjing University of Science and Technology, Nanjing, Jiangsu, China}
\author[2]{Cheng-Zhong Xu}
\ead{czxu@um.edu.mo}
\address[2]{The State Key Laboratory of Internet of Things for Smart City (SKLIOTSC), University of Macau, Macau.}
\author[3]{Le Zhang}
\ead{zhangleuestc@gmail.com}
\address[3]{School of Information and Communication Engineering, University of Electronic Science and Technology of China, Chengdu, Sichuan, China.}
\author[4]{Hui Kong\corref{corl}}
\ead{huikong@um.edu.mo}
\address[4]{Department of Electromechanical Engineering (EME), University of Macau, Macau.}
\cortext[corl]{Corresponding author.}

\address{}

\begin{abstract}
Vision Transformer (ViT) has performed remarkably in various computer vision tasks. Nonetheless, affected by the massive amount of parameters, ViT usually suffers from serious overfitting problems with a relatively limited number of training samples. In addition, ViT generally demands heavy computing resources, which limit its deployment on resource-constrained devices. As a type of model-compression method,  model binarization is potentially a good choice to solve the above problems. Compared with the full-precision one, the model with the binarization method replaces complex tensor multiplication with simple bit-wise binary operations and represents full-precision model parameters and activations with only 1-bit ones, which potentially solves the problem of model size and computational complexity, respectively. In this paper, we investigate a binarized ViT model. Empirically, we observe that the existing binarization technology designed for Convolutional Neural Networks (CNN) cannot migrate well to a ViT's binarization task. We also find that the decline of the accuracy of the binary ViT model is mainly due to the information loss of the \textbf{Attention} module and the \textbf{Value} vector. Therefore, we propose a novel model binarization technique, called \textbf{G}roup \textbf{S}uperposition \textbf{B}inarization (\textbf{GSB}), to deal with these issues. Furthermore, in order to further improve the performance of the binarization model, we have investigated the gradient calculation procedure in the binarization process and derived more proper gradient calculation equations for GSB to reduce the influence of gradient mismatch. Then, the knowledge distillation technique is introduced to alleviate the performance degradation caused by model binarization. Analytically, model binarization can limit the parameter’s search space during parameter updates while training a model. Therefore, the binarization process can actually play an implicit regularization role and help solve the problem of overfitting in the case of insufficient training data. Experiments on three datasets with limited numbers of training samples demonstrate that the proposed GSB model achieves state-of-the-art performance among the binary quantization schemes and exceeds its full-precision counterpart on some indicators. Code and models are available at: \href{https://github.com/IMRL/GSB-Vision-Transformer}{https://github.com/IMRL/GSB-Vision-Transformer}.
\end{abstract}


\end{frontmatter}


\section{Introduction}
In some pattern classification applications, e.g., medical image classification or industrial defect detection, obtaining adequate training samples can be very inconvenient or even impossible. With very limited training samples, modern deep learning techniques, e.g., Visual Transformers, generally cannot achieve good performance due to overfitting problems.
Recently, the transformer-based models ~\citep{dosovitskiy2020image} have been widely explored by researchers. Generally, the continuous improvement of model performance is accompanied by exorbitant computational complexity and excessive storage requirements caused by the enormous amount of parameters. For instance, ViT-L~\citep{dosovitskiy2020image} contains parameters of 300M+ and needs 109G floating point operations (FLOPs), which adversely limit the deployment on resource-constrained devices such as mobile robots and smartphones. Compared with CNN, the transformer lacks the corresponding inductive bias such as locality and spatial invariance. Training with more data is generally required to promote ViT for learning this prior information. Therefore, on relatively small datasets with insufficient training data, the performance of ViT is still unsatisfactory~\citep{dosovitskiy2020image,touvron2021training}. The resulting
overfitting problems can prevent the ViT models from success, although CNN models can also be facing similar situations.

The approaches to address the issue of insufficient training resources mainly focus on four aspects, including data augmentation~\citep{touvron2021training}, network architecture design~\citep{chen2021chasing,wangvtc,lu2021soft,zhou2022training,li2022q,liu2022lgcct,fu2021lmr}, regularization techniques, and training methods that include weak supervision~\citep{wang2023rethinking}, self-supervision, and unsupervised learning.

For network architecture design, researchers have invested tremendous effort to  shrink the model size, such as model compression~\citep{chen2021chasing,wangvtc}, linear attention mechanism~\citep{lu2021soft}, network architecture search~\citep{zhou2022training}, and quantization~\citep{li2022q} et al. Among them, quantization improves the model efficiency by compressing the activations and weights of the model from 32 bits to N ($N<32$) bits without redesigning network architecture through efficient operations ~\citep{li2022q,qin2020forward,lin2020rotated}.

\begin{figure}[htbp]
\setlength{\abovecaptionskip}{5pt}
\setlength{\belowcaptionskip}{0pt}
\centering
\includegraphics[width=3.3in]{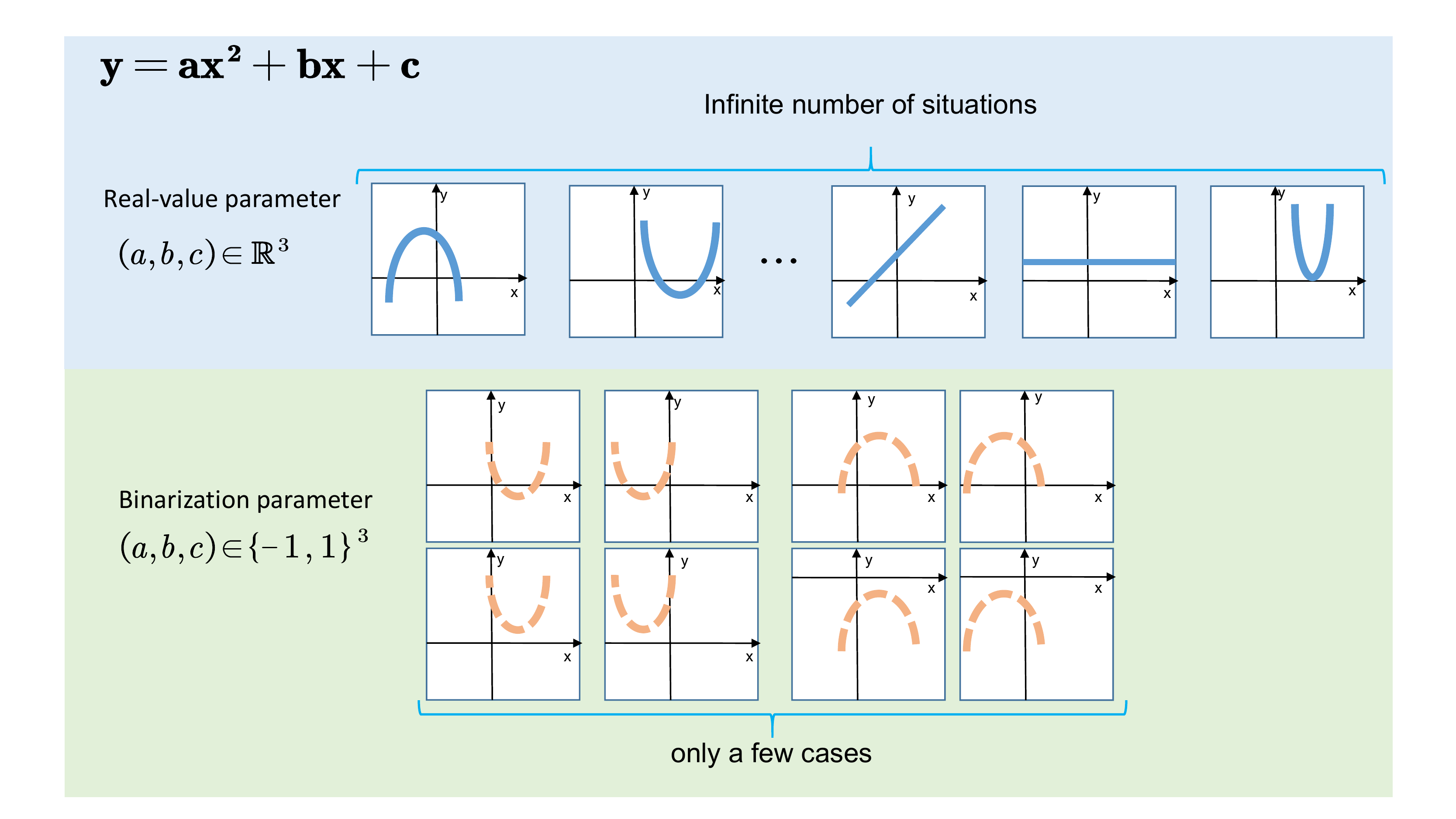}
\caption{An illustrative example to show the effect of model binarization on reducing overfitting problems. This example shows a quadratic model.
We binarize the parameters $a$, $b$, and $c$. Before binarization, the model can potentially take an infinite number of functions. After binarization, the model can only take eight possible functions. Therefore, model binarization can limit the parameter's search space during parameter update while training a model, which can play an implicit regularization role and may help solve the problem of overfitting in the case of insufficient training data ~\citep{santos2022avoiding,ying2019overview,rice2020overfitting,wan2013regularization,srivastava2014dropout}.}
\label{figbir}
\end{figure}

Binarization is an extreme case of quantization, which compresses the weights and activations of neural networks to the 1-bit level. Meanwhile, binarization may be helpful to solve the problem of overfitting. A simple illustration is shown in Fig.~\ref{figbir}. The existing quantitative neural networks can be roughly divided into two categories, Post-Training Quantization (PTQ)~\citep{li2022patch,jeon2022mr} and Quantization-Aware Training (QAT)~\citep{courbariaux2015binaryconnect,rastegari2016xnor,li2022q}. The former directly quantizes the weights on the pre-trained full-precision network, requiring only a small amount of data for calibration or even no retraining at all. Relatively, the latter has completed the model quantization operation in the stage of building the model and trained the quantized model from scratch.
 Inspired by the practicality in CNN-based binary model ~\citep{rastegari2016xnor,liu2020reactnet}, researchers dedicated to exploiting the binary transformer in the field of \textbf{N}atural \textbf{L}anguage \textbf{P}rocessing(NLP)~\citep{qin2022bibert,liu2022bit} and achieved promising performance. However, as mentioned  in~\citep{qin2023bibench}, the binary transformer has a larger binarization error than CNN-based or \textbf{M}ulti-\textbf{L}ayer \textbf{P}erceptrons (MLP)-based models.

In this paper, we have made innovations and improvements for the above-mentioned problems. We propose an efficient ViT binarization method, which solves the problems of full-precision ViT such as high computational complexity and unable to achieve good performance when trained on small datasets. Our contributions are summarized in the following aspects.

\indent $\bullet$ Through comparative experiments, we found that the binarization error of the $Attention$ module is one important factor that decreases the binary ViT performance. Accordingly, we proposed the GSB binarization to group the input vectors, i.e. attention and vector value, by their values and then superimpose them after separate binarization.

\indent $\bullet$ Due to the non-differentiability of the binarization function, the proposed binary scheme needs to complete backpropagation through subgradient techniques. Therefore, we derive the proper gradient calculation for GSB based on \textbf{S}traight \textbf{T}hrough \textbf{E}stimator (STE)~\citep{hinton2012neural,bengio2013estimating} to prevent training failure caused by gradient mismatch.

\indent $\bullet$ ViT is vulnerable to overfitting with insufficient data. However, the ViT based on GSB binarization can reduce the effect caused by overfitting. Regarding this phenomenon, we rethink the significance of binarization error in the training process and give an explanation for this phenomenon.

\indent $\bullet$ Compared with the CNN-based and MLP-based models, the performance degradation caused by binarization is more serious in the transformer models. We infer the cause of this phenomenon and explain it in the paper.\\
\indent The remainder of this paper is as follows. In Section~\ref{section2}, related works on model binarization and training ViT with limited training samples are summarized. In Section ~\ref{section3}, we propose a baseline binarization ViT model. In Section~\ref{section4}, our GSB ViT model is introduced in detail.
In Section~\ref{section5}, we evaluate the performance of the proposed method on small datasets where the amount of data is insufficient or the training labels are noisy. Finally, conclusions are presented in Section~\ref{section6}.
\section{Related Works}\label{section2}
 \noindent \textbf{Quantization and Binarization.} Courbariaux et al.~\citep{courbariaux2015binaryconnect} was the first to propose QAT-based binary Neural Networks (BNN), a model with binary weights and full-precision activations. Meanwhile, different from the straight-through estimator (STE) proposed by Hinton~\citep{hinton2012neural}, the $hard$-$tanh$ function was utilized to replace the gradient of the quantization function. Compared to just binarizing the weights, binarizing activations simultaneously further promotes the improvement of the compression proportion of the model. However, further binarization brings about a sharp decline in performance. Rastegari et al.~\citep{rastegari2016xnor} solved this problem. Meanwhile, the approach, bitwise operations, only applicable to binarized networks was proposed to approximate convolutions and maximize the acceleration effect of the model. In recent years, the research direction of model binarization mainly focuses on several aspects such as the scheme of quantization~\citep{Liu_2022_CVPR} and gradient fitting of quantization function~\citep{liu2018bi,xu2022recurrent}. To solve the problem of reduced representational ability of the model caused by the binarization of activation, Liu et al.~\citep{liu2018bi} proposed a shortcut connection between the sign function and batch normalization to mitigate the performance degradation of the model. Martinez et al.~\citep{martinez2020training} utilized the intermediate output of the full-precision network to supervise the corresponding binarized network, which confirmed that aligning the distribution of the corresponding output is beneficial to improve the performance of the binarization network. Xu et al.~\citep{xu2022recurrent} proposed recurrent bilinear optimization to improve the performance of BNN. Inspired by~\citep{martinez2020training},  Liu et al.~\citep{liu2020reactnet} aligned the output distribution of the full-precision network with the corresponding binary network by reshaping and shifting the activation distribution. Xu et al.~\citep{xu2021learning} applied the Fourier transform to fit the gradient of the sign function in the frequency domain, alleviating the gradient mismatch caused by STE. Lin et al.~\citep{lin2017towards} applied a linear combination of binary weight bases and binary activation bases to fit the full-precision weight and activation, respectively. Different from our work, we only perform group superposition binarization on the activation part, which will reduce the computational complexity of the forward process compared to ABC-Net~\citep{lin2017towards}. Meanwhile, our idea of superposition is different from ABC-Net. Our method manages to reserve the ability of binarized activation to highlight the difference between the different values of the corresponding full-precision one. Therefore, we only binarize the full-precision values once, and the rest of the binarized components are represented by the masking operations by thresholding comparisons. On the contrary, the objective of the multiple activation base proposed in ABC-Net is to enhance the representation ability of activation, which does not highlight the difference between different values in the corresponding full-precision one. Zhuang et al.~\citep{zhuang2019structured} proposed Group-Wise Binary Decomposition, which applies multiple binarized convolutional layers to fit the convolutional layer with full-precision weights. Meanwhile, a strategy for automatic decomposition based on the gating mechanism was proposed to help the model select the suitable number of binary convolutional layers based on the activations. Unlike our work, Group-Net~\citep{zhuang2019structured} focuses on combinations of binarized weights. Since the attention module cannot be decomposed into a superposition of multiple binary layers, this method cannot be well applied to the task of binarization of the Vision Transformer.  \\
    \indent The work about binarizing the transformer is mainly concentrated in NLP. The corresponding work of CV mainly focuses on low-bit quantification. Qin et al.~\citep{qin2022bibert} considered the distribution of the attention value and the characteristics of the softmax function, and introduced the bool function to set the binary state of the attention at 0 and 1. Liu et al.~\citep{liu2022bit} analyzed the characteristics of different modules of the transformer, proposed two different quantization schemes, and studied the corresponding gradient propagation process. Li et al.~\citep{li2022q} maximized the information entropy of vector Q and vector K through layer normalization, which improves the information representation of the quantized attention module. Then, the distribution consistency of the full-precision attention and the corresponding quantization module was improved by utilizing distillation learning. He et al.~\citep{he2022bivit} improved the attention quantization method proposed by~\citep{qin2022bibert} to realize partial binarization of ViT. In order to ensure the performance of the model, the activation of the MLP module was not binarized.\\
\hspace*{\fill} \\
\textbf{Training ViT on a small dataset from scratch.} Very few works have investigated how to train ViT on small datasets~\citep{liu2021efficient,chen2022improving,lu2022bridging,chen2023accumulated,cao2022training,li2022locality}. Compared with CNN, ViT lacks the former's unique inductive bias, and more data are required to force ViT to learn this prior. Therefore, training ViT with insufficient data is an urgent problem that needs to be solved. With a unique data enhancement technology and distillation, a pre-trained CNN was applied as a teacher model by Touvron et al.~\citep{touvron2021training} to train a ViT model without massive data. Liu et al.~\citep{liu2021efficient} introduced a self-supervised auxiliary task that learns relative spatial information to improve the performance of ViT trained from small datasets. Chen et al.~\citep{chen2022improving} combined frequency domain information to the feature extractor of ViT for improving the performance of the model. By introducing convolution operations and using the information of each patch to correct class tokens, Lu et al.~\citep{lu2022bridging} improved the spatial correlation and channel discrimination of ViT. Chen et al.~\citep{chen2023accumulated} suppressed the noise effect caused by weak attention values and improved the performance of ViT on small datasets. Cao et al.~\citep{cao2022training} proposed a method called parametric instance discrimination to construct the contrastive loss to improve the feature extraction performance of ViT trained on small datasets. Li et al.~\citep{li2022locality} applied a CNN-based teacher model to guide ViT to improve the ability about capturing local information.

\section{The baseline of binarization method}\label{section3}
In this section, we introduce our baseline model in very detail before presenting our GSB binarization method. First, in subsection~\ref{section3_1}, we introduce the binarization process of weights and activations in the linear layer based on the binarization method proposed by XNOR-Net~\citep{rastegari2016xnor} and the classification method proposed by BiT~\citep{liu2022bit} for the distribution of activation. Subsequently, in subsection~\ref{section3_2}, we propose the improvement of the binarization process of the attention module.
\begin{figure*}[htbp]
\setlength{\abovecaptionskip}{5pt}
\setlength{\belowcaptionskip}{0pt}
\centering
\includegraphics[width=6.5in]{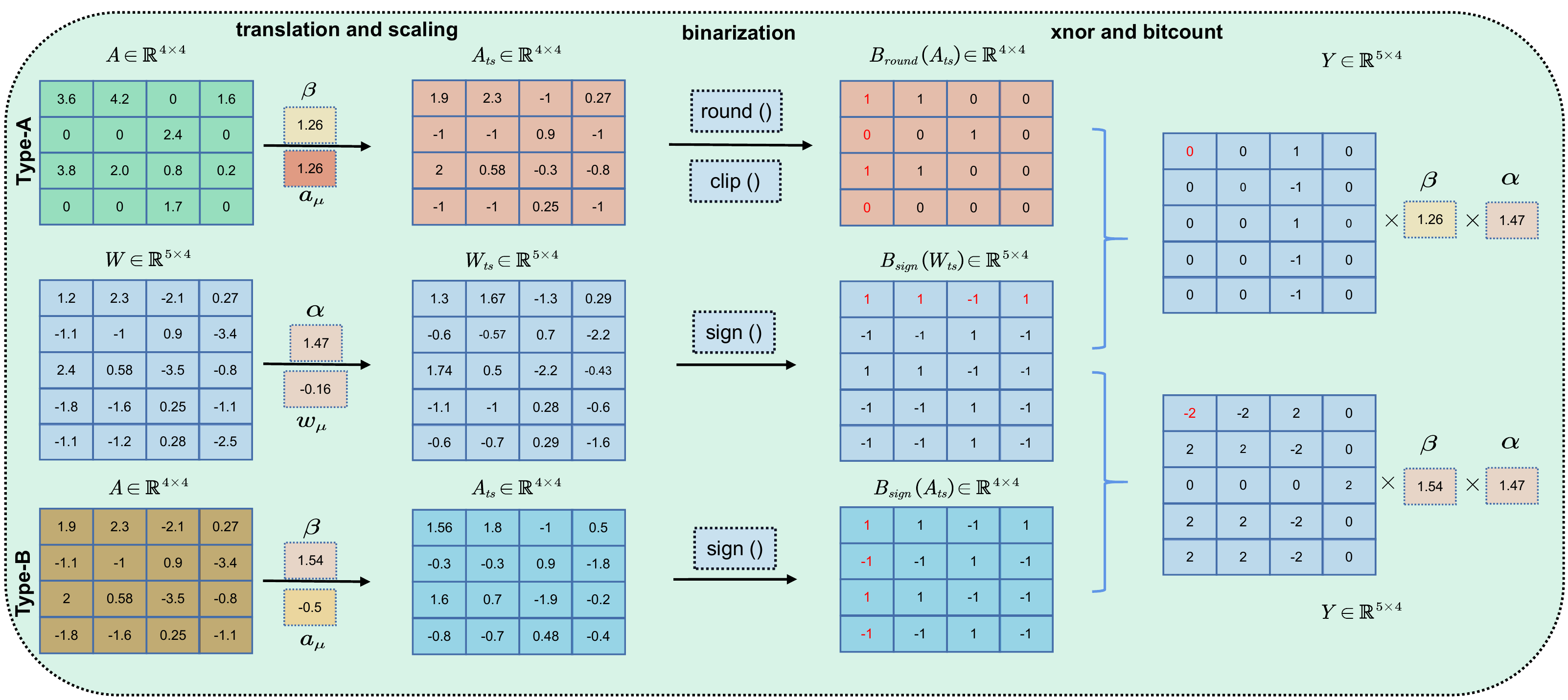}
\caption{The binarization process of the linear layer. $\mathbf{Y}\in \mathbb{R}^{d\times n}$ indicates the output of the linear layer. $\mathbf{W}\in \mathbb{R}^{d\times b}$ and $\mathbf{A}\in \mathbb{R}^{b\times n}$ refer to the full-precision weight and the activation, respectively.  We take $n=4$, $d=5$, and $b=4$ as an example. The batch size is 1. The $xnor$ and $popcount$ operations with 0-valued vectors can refer to the work~\citep{qin2022bibert}.}
\label{figlinear}
\end{figure*}
\begin{figure}[!t]
\setlength{\abovecaptionskip}{5pt}
\setlength{\belowcaptionskip}{0pt}
\centering
\includegraphics[width=2.5in]{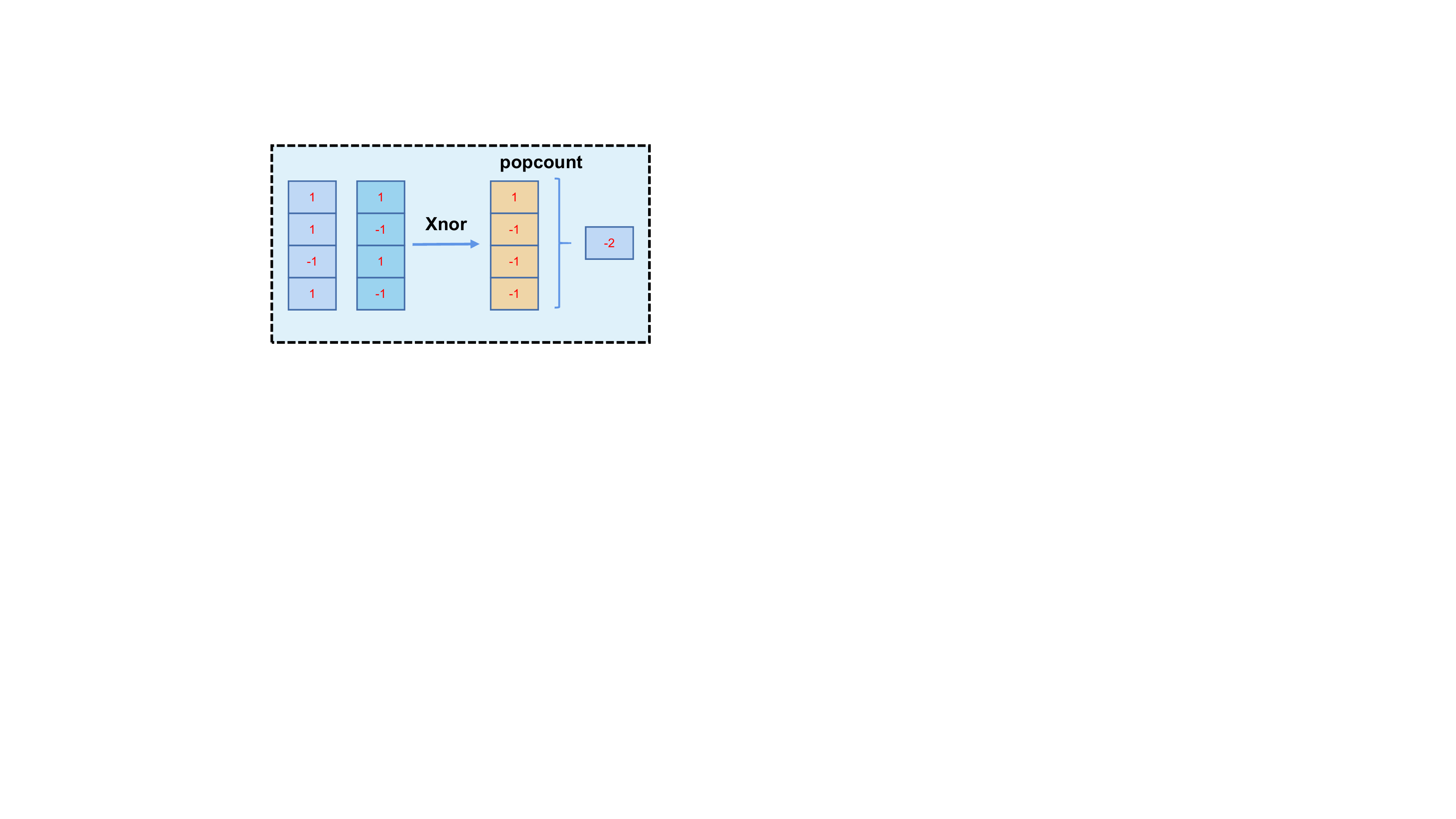}
\caption{The bitwise $xnor$ and $popcount$ operation.}
\label{figxnor}
\end{figure}
\subsection{The binarization of Linear layer} \label{section3_1}

The linear layer exists in the entire process of the transformer. Its binarization process can be described as Eq.~\ref{eq1} and shown in Fig.~\ref{figlinear}.
\begin{equation}
\label{eq1}
\begin{aligned}
	\mathbf{Y}=\mathbf{W}*\mathbf{A}\approx \alpha \beta \left( B\left( \mathbf{W}_{ts} \right) \circledast  B\left( \mathbf{A}_{ts} \right) \right) ,\\
\end{aligned}
\end{equation}
where $\mathbf{Y}\in \mathbb{R}^{d\times n}$ indicates the output of the linear layer. $\mathbf{W}\in \mathbb{R}^{d\times b}$ and $\mathbf{A}\in \mathbb{R}^{b\times n}$ refer to the full-precision weight and the activation, respectively. $n$ represents the feature index, and $b$ and $d$ denoting the channel index.
$\mathbf{W}_{ts}\in \mathbb{R}^{d\times b}$ and $\mathbf{A}_{ts}\in \mathbb{R}^{b\times n}$ are the normalized $\mathbf{W}$ and $\mathbf{A}$ based on translation and scaling, respectively. Specifically, $\mathbf{W}_{ts}=\left( \frac{\mathbf{W}-w_{\mu}}{\alpha} \right)$ and $ \mathbf{A}_{ts}=\left( \frac{\mathbf{A}-a_{\mu}}{\beta} \right)$. $B()$ is the corresponding binarization function. $*$ means matrix multiplication and $\circledast $ denotes the binary matrix multiplication by bitwise $xnor$ and $popcount$ operation as shown in Fig.~\ref{figxnor}.
The $\alpha$ and $\beta$ represent the scaling factors. $w_{\mu}\in \mathbb{R}^{1\times 1}$ and $a_{\mu}\in \mathbb{R}^{1\times 1}$ are the mean value of $\mathbf{W}$ and $\mathbf{A}$, respectively. Following XNOR-Net~\citep{rastegari2016xnor}, the detailed calculations are listed in Eq.~\ref{eq1-2}.
\begin{equation}
\label{eq1-2}
\begin{aligned}
w_{\mu}&=\frac{1}{n_W}\sum_{i=1}^b{\sum_{j=1}^d{w_{ij}}},a_{\mu}=\frac{1}{n_A}\sum_{i=1}^n{\sum_{j=1}^b{a_{ij}}},\\
\alpha &=\frac{1}{n_W}\left\| \mathbf{W} \right\| _{l_1},\beta =\frac{1}{n_A}\left\| \mathbf{A} \right\| _{l_1},\\
\left\| \mathbf{W} \right\| _{l_1}&=\sum_{i=1}^b{\sum_{j=1}^d{\left| w_{ij} \right|}},\left\| \mathbf{A} \right\| _{l_1}=\sum_{i=1}^n{\sum_{j=1}^b{\left| a_{ij} \right|}}, \\
\end{aligned}
\end{equation}
where $\left\| \mathbf{W} \right\| _{l_1}\in \mathbb{R}^{1\times 1}$ and $\left\| \mathbf{A} \right\| _{l_1}\in \mathbb{R}^{1\times 1}$ are the $l_1$ norms of $\mathbf{W}$ and $\mathbf{A}$, respectively. $n_W$ and $n_A$ are the number of elements of $\mathbf{W}$ and $\mathbf{A}$ respectively. $w_{ij}$ and $a_{ij}$ are the element of $\mathbf{W}$ and $\mathbf{A}$, respectively. \\
\indent In BiT~\citep{liu2022bit}, the elements in the activation $\mathbf{A}$ of the Transformer have two different types of distributions. One type of activation only contains non-negative elements, such as the attentions and vectors following the ReLU function. This kind of activation is denoted as \textit{type-A} activation. The other activation type contains positive, negative, and zero elements, denoted as \textit{type-B} activation.  \\
\indent To reduce information loss, we follow BiT~\citep{liu2022bit} by adopting two different binarization schemes with respect to the two types of activations. Note that using the $sign$ function to complete the binarization process of the attention matrix will get an all-one matrix and destroy the effect of the attention module. As shown in Eq.~\ref{eq2}, we utilize the $round$ and $clip$ functions to binarize \textit{type-A} activation.  Meanwhile, Eq.~\ref{eq2-1} is applied to accomplish the binarization of \textit{type-B} activation, where we follow the XNOR-Net~\citep{rastegari2016xnor} to apply the $sign$ function for binarizing the weights of ViT.
\begin{equation}
\label{eq2}
\begin{aligned}
B_{round}\left( x^{i,j} \right)&=clip\left( round\left( x^{i,j} \right) ,0,1 \right)=\begin{cases}
	0&		x^{i,j}\leqslant 0.5\\
	1&		x^{i,j}>0.5\\
\end{cases},
\end{aligned}
\end{equation}
where $x^{i,j}$ is the element of the input $\mathbf{X}$. $clip(x,l,h)$ refers to compressing the input value $x$ into the interval $[l, h]$, and the $round$ function indicates rounding the input value to the nearest integer.\\
\begin{equation}
\label{eq2-1}
B_{sign}\left( x^{i,j} \right)=sign\left( x^{i,j} \right) =\begin{cases}
	1&		x^{i,j}\geqslant 0\\
	-1&		x^{i,j}<0\\
\end{cases}.
\end{equation}
\\
\indent The non-differentiability of $sign$, $round$, and $clip$ functions will make the model unable to be trained with back-propagation. To solve this problem, the subgradient method is regularly utilized, e.g., the gradient calculation of the $ReLU$ function. We follow the theory proposed by Hinton~\citep{hinton2012neural} and apply STE proposed in~\citep{bengio2013estimating}, as shown in Eq.~\ref{eq3}, to handle the gradient calculation of $sign$, $round$, and $clip$ functions.
\begin{equation}
\label{eq3}
\begin{aligned}
G_{sign\left( x^{i,j} \right)}&=\begin{cases}
	1&		\left| x^{i,j} \right|\leqslant 1\\
	0&		otherwise\\
\end{cases},
\\
G_{round\left( x^{i,j} \right)}&=\begin{cases}
	1&		x^{i,j}\notin \mathbb{Z}\\
	0&		otherwise\\
\end{cases},
\\
G_{clip\left( x^{i,j},l,h \right)}&=\begin{cases}
	1&		l\leqslant x^{i,j}\leqslant h\\
	0&		otherwise\\
\end{cases}.
\end{aligned}
\end{equation}
\indent For calculation of the gradient in the binarization process of weight, due to the existence of outliers in the weights, only updating the weights in the range [-1,1] will make the outliers never be updated, which will eventually affect the performance of the model. This problem is also mentioned in \citep{liu2022bit,xu2021recu}. On the contrary, the activation value does not need to consider this problem during its binarization process because it is always changing. Therefore, we apply Eq.~\ref{eqw} to define the gradient of the process of weight binarization.
\begin{equation}
\label{eqw}
\begin{aligned}
G_{sign\left( x^{i,j} \right)}^{weight}&=1.
\end{aligned}
\end{equation}

\subsection{The binarization of ViT's attention module} \label{section3_2}
With little adaption, straightforwardly, one way of binarization of the attention matrix can be implemented as in Eq.~\ref{eqatt}.
\begin{equation}
\label{eqatt}
\begin{aligned}
&\mathbf{A_{re}}=softmax\left( \frac{B\left( \mathbf{Q} \right) *B\left( \mathbf{K} \right)}{\sqrt{d}} \right) ,
\\
&\mathbf{A_{bin}}=clip\left( round\left( \frac{\mathbf{A_{re}}}{\gamma} \right),0,1 \right) ,
\end{aligned}
\end{equation}
where $\mathbf{Q}$ and $\mathbf{K}$ represent the query and key matrices in the ViT, respectively. $d$ is the number of channels for $\mathbf{Q}$ and $\mathbf{K}$. $B()$ is the binarization function defined in Eq~\ref{eq2-1}. $softmax$ is the softmax function. $\gamma$ is a scale factor, which is computed in the same way as in Eq.~\ref{eq1-2}. $\mathbf{A_{re}}$ is a real-valued attention matrix.

\begin{figure}[htbp]
\setlength{\abovecaptionskip}{5pt}
\setlength{\belowcaptionskip}{0pt}
\centering
\includegraphics[width=6.0in]{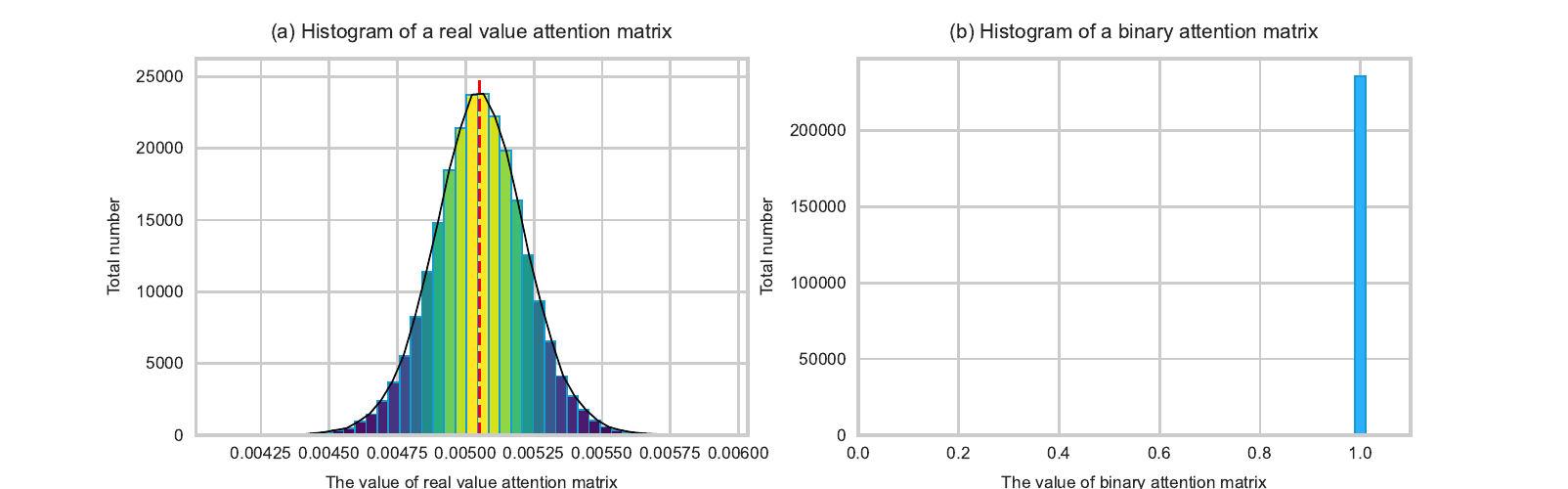}
\caption{The distribution of the values of the attention matrix closest to the output layer (the layer that outputs class probabilities). The model utilized is our baseline at the initialization stage. (a) The distribution of the values of $\mathbf{A_{re}}$. (b) The distribution of the values of the binarized attention matrix that has been scaled by the scaling factor defined by Eq.\ref{eq1-2}.}
\label{figbasebi}
\end{figure}

Different from the binarization of the activation in the linear layer (Section \ref{section3_1}), the softmax operation has transformed the attention values of ViT to the interval of $[0, 1]$. Therefore, the translation operations can be skipped during the binarization of the attention module.
Empirically, we have observed that the binarization of $\mathbf{Q}$ and $\mathbf{K}$ and the scaling factor $\sqrt{d}$ have made almost all the values of $\mathbf{A_{re}}$ smaller than 0.5 and very concentrated at initialization stage, which is shown in Fig.~\ref{figbasebi} (a). Thereafter, as shown in Fig.~\ref{figbasebi} (b), the scaling operation in the $clip$ function (division by $\gamma$) turns almost all values into 1. In this case, the scaling operation destroys the effect of the attention module, i.e., information can interact between patches across the whole image. On the contrary, removing the scaling operation and directly binarizing the attention matrix will result in an all-zero matrix $\mathbf{A}_{bin}$, which also corrupts the information interaction in the attention module and will directly lead to training failure.\\
\indent The purpose that applying the scaling operation to the activation is to reduce the difference between the distribution of binarized value and the corresponding full-precision one. However, we observe that using the scaling factor defined by Eq.\ref{eq1-2} will have the opposite effect instead when applied to the attention matrix, i.e., it increases this difference because too many attention values close to 0 are binarized to 1.

Although using the factor defined by Eq.\ref{eq1-2} will introduce large binarization errors and the inability to highlight which patches are more important, it can help the model update the weight to process the extreme situation (shown in Fig.~\ref{figbasebi}) in the distribution of attention values. With this consideration, we train the model in two stages. In the first training stage, we apply the scale factor defined in Eq.\ref{eq1-2} to scale $\mathbf{A_{re}}$. After the model converges in this stage, Eq.\ref{eq1-3} is adopted to redefine the scale factor in the second training stage to reduce binarization error.

There is a problem that needs to be considered. When the values of  $\mathbf{A_{re}}$ are all less than 0.5, applying the scale factor defined by Eq.\ref{eq1-3} will cause the failure of the training procedure. Therefore, we introduce a switch mechanism for the training of our baseline. When the number of values (larger than the binarization threshold) in the scaled $\mathbf{A_{re}}$ (scaled by Eq.\ref{eq1-3}) is smaller than the number of image patches, the scaling operation is re-implemented by Eq.\ref{eq1-2} to avoid the training failure.

\begin{equation}
\label{eq1-3}
\begin{aligned}
\gamma =\frac{1}{n_{W_m}}\left\| \mathbf{a}_m \right\| _{l_1}, \left\| \mathbf{a}_m \right\| _{l_1}=\sum_i^{n_{W_m}}{\left| a_{m}^{i} \right|},
\end{aligned}
\end{equation}
where $n_{W_m}$ represents the number of attention values larger than the binarization threshold (0.5), $a_{m}^{i}$ is an attention value larger than 0.5. $\mathbf{a}_m$ indicates the vector consisting of all attention values that are larger than the binarization threshold.

\section{Group Superposition Binarization}\label{section4}
\begin{figure}[htbp]
\setlength{\abovecaptionskip}{5pt}
\setlength{\belowcaptionskip}{0pt}
\centering
\includegraphics[width=3.5in]{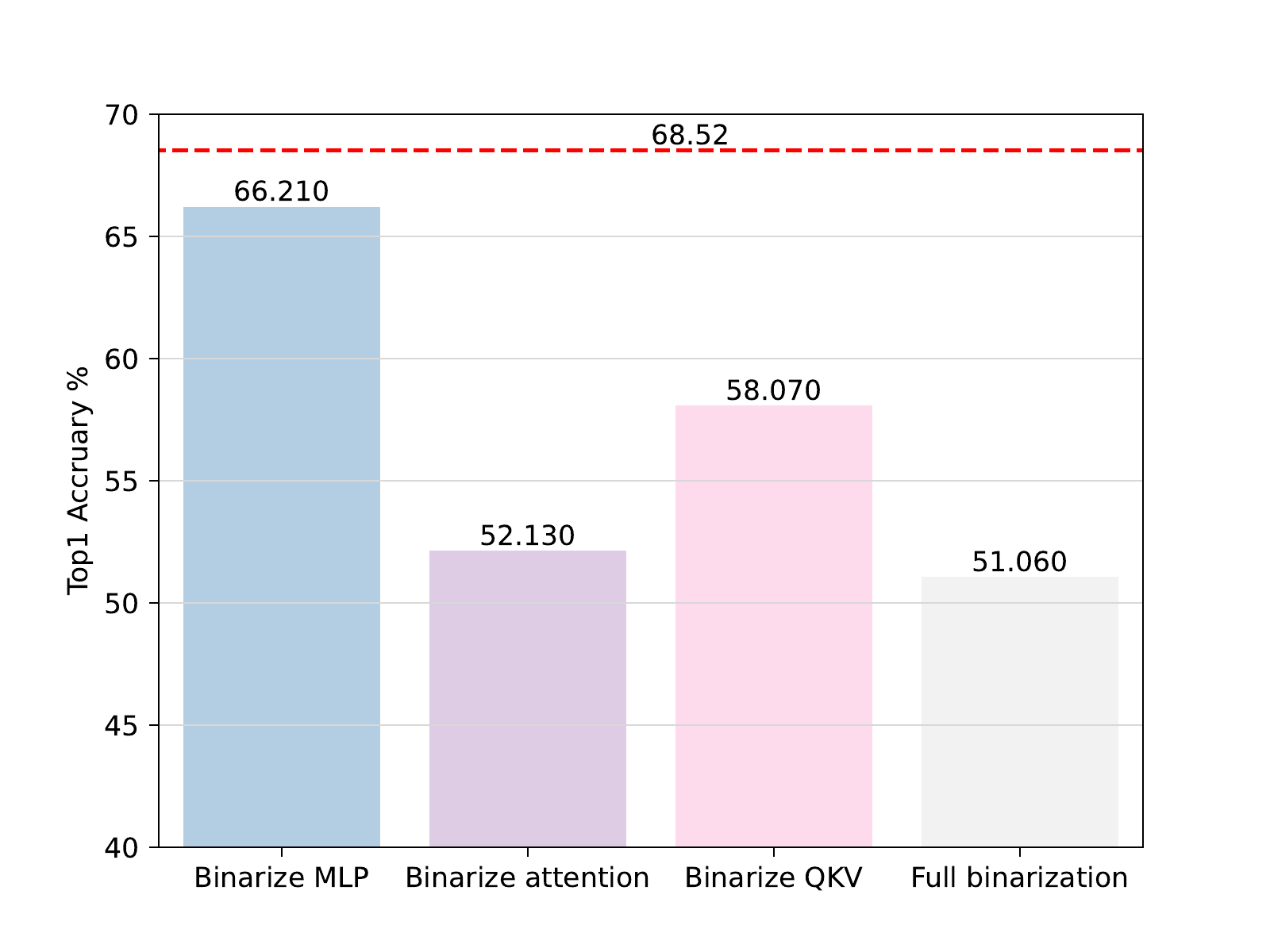}
\caption{The performance comparison of different modules of ViT after binarization on the CIFAR-100 dataset. The performance of the full-precision model is represented by the horizontal dashed line. The performance of all compared models is obtained by training 600 epochs from scratch on the dataset. The base model is DeiT-small~\citep{touvron2021training} without distillation, which can be seen as a ViT with strong data augmentation.}
\label{fig1}
\end{figure}

Although our baseline has been designed in a fine manner, the performance gap between the corresponding full-precision network and the binary one is still large. Inspired by Q-ViT~\citep{li2022q}, we show a detailed comparison of the binarization performance of different modules of ViT. As shown in Fig.~\ref{fig1}, the binarization of the attention values, $\mathbf{Q}$, $\mathbf{K}$, and $\mathbf{V}$ matrices have a large impact on the performance of the model. In the following section, we give our analysis to answer why our baseline model cannot obtain satisfying performance.
\subsection{Why binarized ViT is weaker than binarized CNN and binarized MLP?}
\begin{figure*}[htbp]
\setlength{\abovecaptionskip}{5pt}
\setlength{\belowcaptionskip}{0pt}
\centering
\includegraphics[width=6.5in]{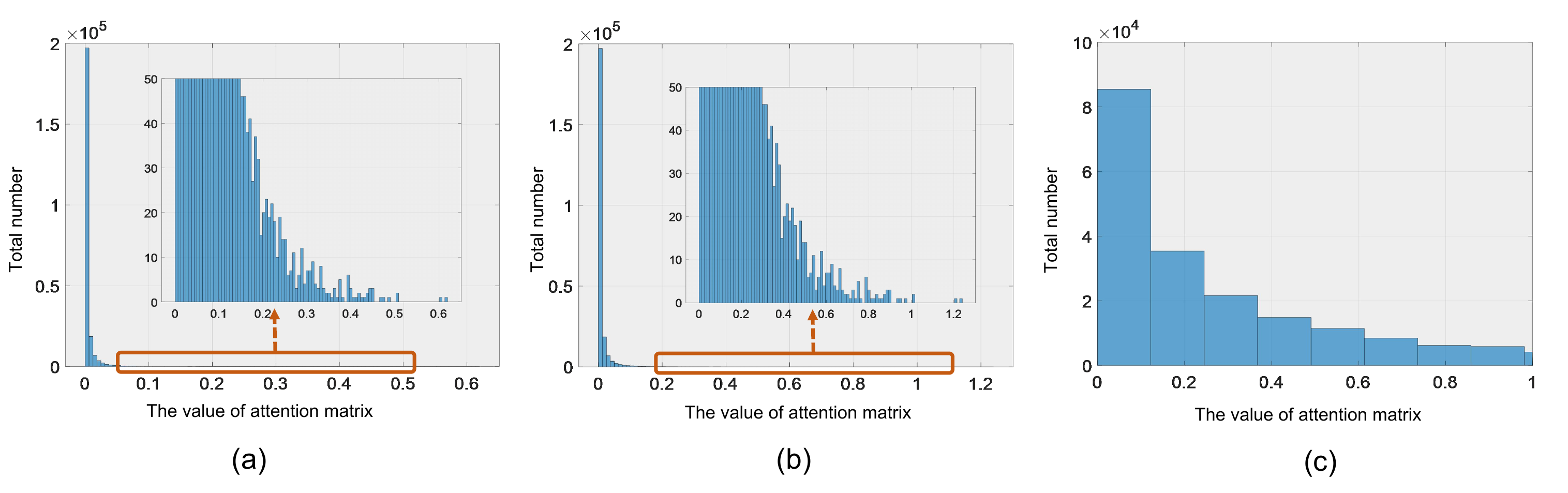}
\caption{The distribution of the values of the attention layer (the one closest to the output layer). The adopted model is our baseline trained on the CIFAR-100 dataset. (a) The distribution of the values of $\mathbf{A_{re}}$. (b) The distribution of the values of scaled $\mathbf{A_{re}}$ by Eq.~\ref{eq1-3}. (c) The distribution of the values of scaled $\mathbf{A_{re}}$ by Eq.~\ref{eq1-2}. The red dotted box represents the partially enlarged figure, and the black vertical line represents the binarization threshold.}
\label{fig2}
\end{figure*}

As a typical example, we show the distribution of the attention values of $\mathbf{A_{re}}$ for a typical model trained on the CIFAR-100 in Fig.~\ref{fig2} (a).
The scaled $\mathbf{A_{re}}$ by Eq.~\ref{eq1-3} and Eq.~\ref{eq1-2} is shown in Fig.~\ref{fig2} (b) and (c), respectively. Intuitively, the values of scaled $\mathbf{A_{re}}$ by Eq.~\ref{eq1-3} and Eq.~\ref{eq1-2} follow a long-tailed distribution, which is also mentioned in BiViT~\citep{he2022bivit}.


\begin{figure*}[htbp]
\setlength{\abovecaptionskip}{5pt}
\setlength{\belowcaptionskip}{0pt}
\centering
\includegraphics[width=5.2in]{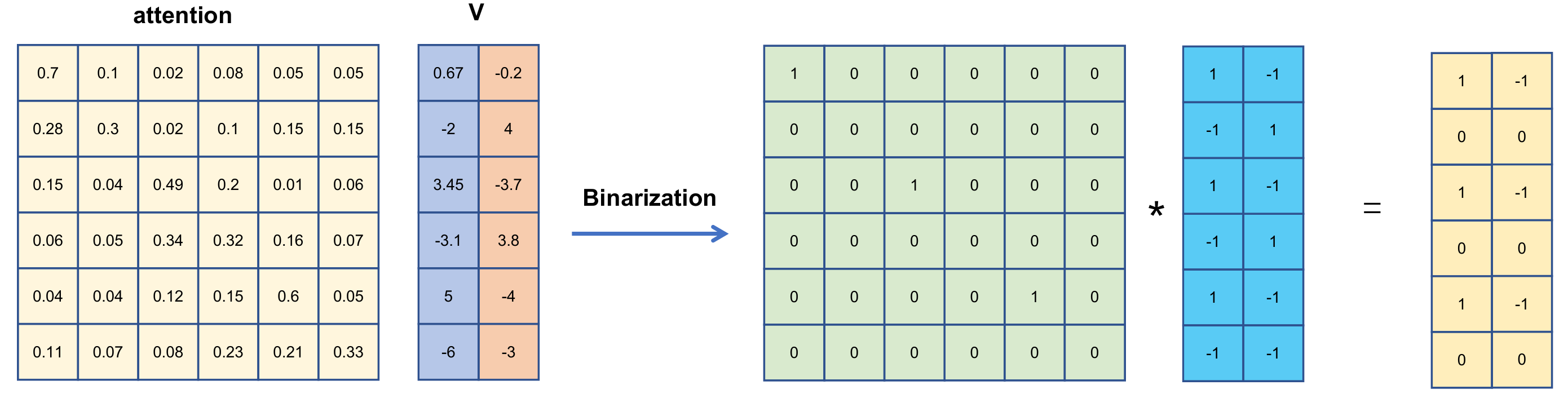}
\caption{Vanishing feature problem caused by binarized attention matrix. We take $N=6,C=2$ as an example of $V$. When a row in the attention matrix is all 0 after binarization, the information of all channels of the corresponding matrix V will be set to 0.}
\label{fig3}
\end{figure*}
 \indent  The attention matrix assists the information interaction between image patches in the transformer. Moreover, the essence of self-attention is the similarity between features of image patches. Similarity value refers to the importance of the image patch in the information interaction with the target image patch. For the binarization method with the scale factor defined in Eq.~\ref{eq1-2}, too many values close to 0 in the $\mathbf{A}_{re}$ are set to 1. In this case, the difference between the similarity values corresponding to all image patches will be reduced, which will introduce excessive error in the information interaction phase and cause noticeable performance degradation of the model. For the binarization method with the scale factor defined in Eq.~\ref{eq1-3}, almost 99\% of the attention values will be rounded to 0, which will result in extreme sparsification of the attention matrix. Therefore, a serious problem will arise when only a few values in the scaled $\mathbf{A}_{re}$ are larger than the binarization threshold (0.5). An example is shown in Fig.~\ref{fig2} (b), where the number of scaled attention values larger than the binarization threshold is less than the number of tokens about image patch of our model (same as DeiT-small~\cite{touvron2021training} has 196 tokens of image patch and two additional class tokens). In this case, as shown in Fig.~\ref{fig3}, the feature of some patches will be erased by the attention mechanism. In particular, The final output of the model will be zero when the feature of the class token is eliminated. This is the problem of vanishing features, which often occurs when there are multiple similar patches in the image. Researchers often apply positional embedding to make ViT have the ability to distinguish similar patches in images. However, the binarization operation will implicitly weaken the effect of positional embedding. \\
\begin{figure*}[htbp]
\setlength{\abovecaptionskip}{5pt}
\setlength{\belowcaptionskip}{0pt}
\centering
\includegraphics[width=5.0in]{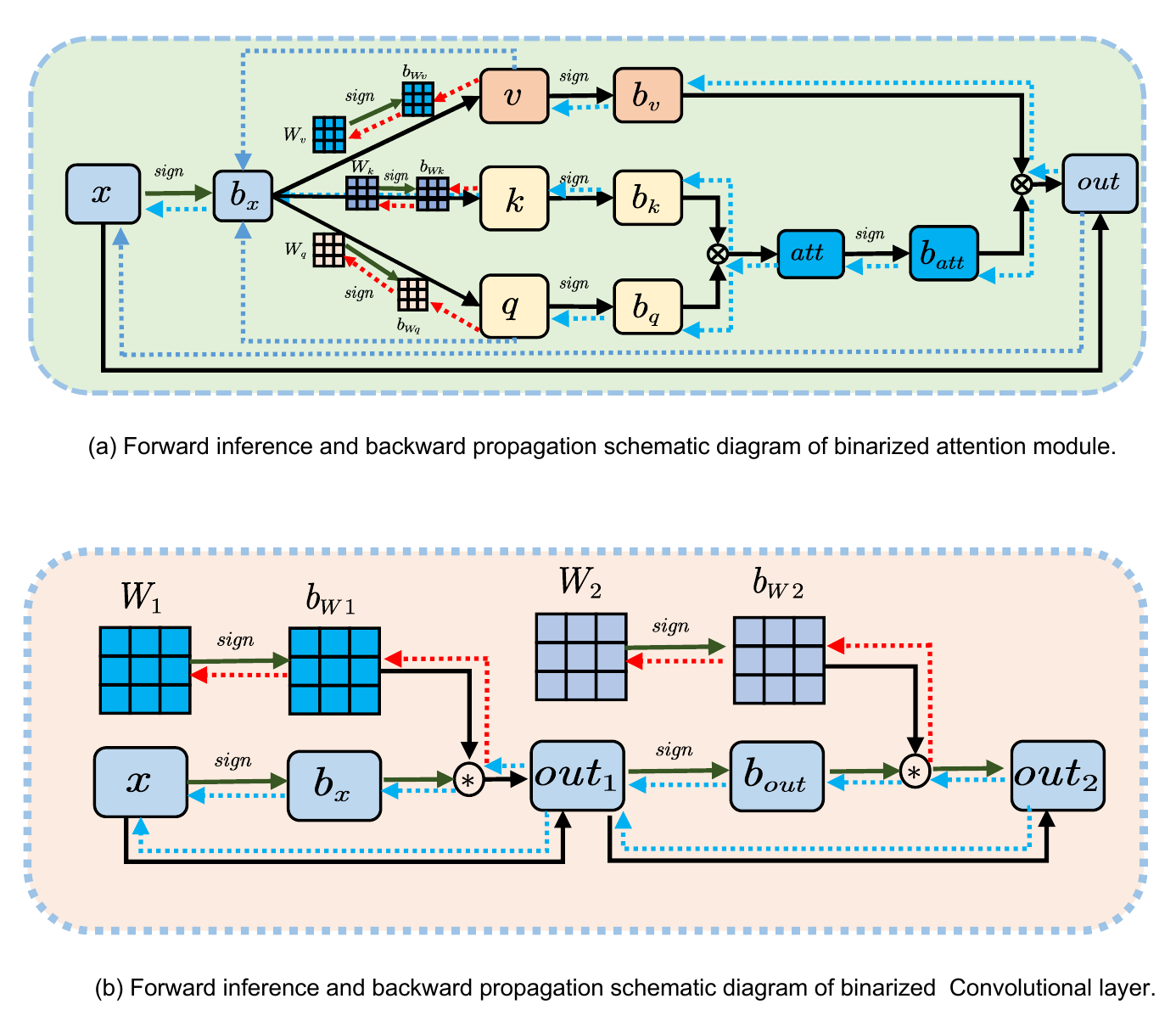}
\caption{Forward inference and backward propagation schematic diagram of binarized attention module and convolutional layer.The black arrows represent the forward propagation path of the model, the red arrows represent the gradient propagation path of the weight parameters, and the blue arrows represent the gradient propagation path of the activations.}
\label{figfbac}
\end{figure*}
\indent For the forward propagation of the model, in DeiT~\cite{touvron2021training}, the class token serves as the final feature representation. Compared to the global pooling in CNN, the update of the class token relies on the information interaction among tokens at the token level through the attention mechanisms at different stages to learn the features in the image. Therefore, the accuracy of the attention matrix affects the feature representation performance of the class token. Due to the properties of the softmax function, the binarized attention matrix exhibits an asymmetric distribution. Moreover, based on the observation in Fig.~\ref{fig2} (a), it can be inferred that the binarized attention matrix is sparse, as only a small fraction of image tokens interact with the class token within an MHA (multi-head attention) module. Consequently, the binarized VIT only includes a small portion of image information as the final feature. On the other hand, the binarized CNN's convolutional modules expand receptive fields by stacking convolutional layers repeatedly, the application of global pooling enabling the aggregation of global image features. Similarly, the binarized MLP model can also aggregate global image features. Due to the characteristics of convolutional operations, the binarization of the Convolutional layer in CNN can apply the technique that channel-wise scale factors~\cite{qin2023bibench} to enhance weights and activate search scope without increasing computational complexity. In contrast, this technique cannot be applied to the calculation process of the attention matrix and the FC layer. Therefore, without increasing extra computational cost, the representational capacity of binarized ViT is significantly lower than that of the CNN model.\\
\indent Regarding the backward propagation of the model, as shown in Fig~\ref{figfbac},  the attention module involves two sign functions between the V weight matrix and the final output of the module (For the sake of simplicity, we will ignore the fully connected layer in the attention module.), while Q and K entail three sign functions due to the binarization of the attention matrix. In comparison, CNN only has one sign function. As a result, the gradient error in the backward propagation of the attention module will be much larger than that of the CNN model. Additionally, the class token is essentially a group of weight parameters added at the beginning of the model. Although residual connections can propagate some precise gradients, the weight on the class token still suffers from significant gradient errors. Meanwhile, The gradient characteristics of the softmax function itself and the truncation properties of the binarized activation gradient make the Q and K parameter matrices prone to gradient vanishing. These factors contribute to the greater difficulty in optimizing the binarized ViT model compared to CNN and MLP models.
\subsection{Attention mechanism based on group superposition binarization}
 \begin{figure*}[htbp]
\setlength{\abovecaptionskip}{5pt}
\setlength{\belowcaptionskip}{0pt}
\centering
\includegraphics[width=6.0in]{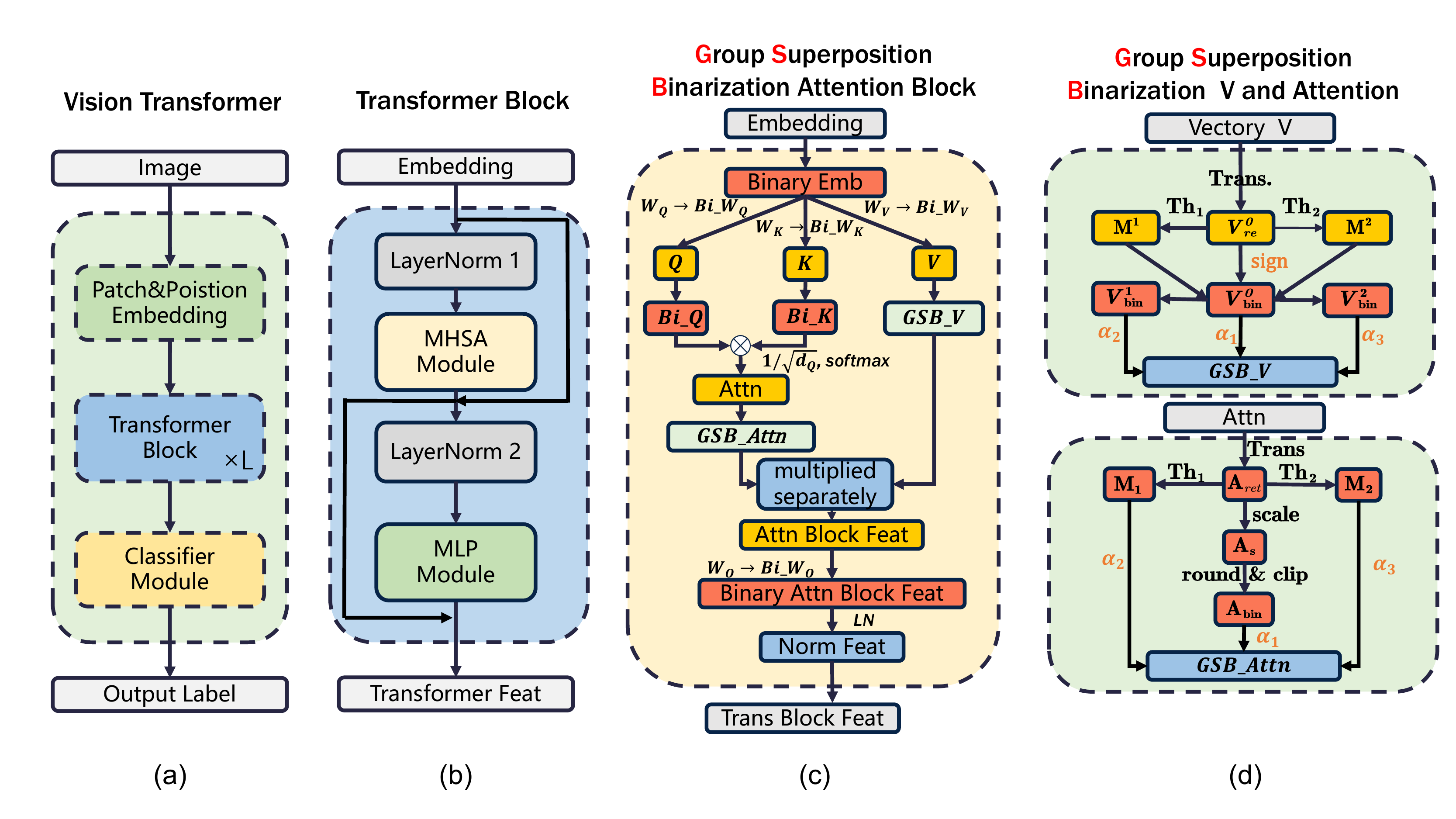}
\caption{The framework of our proposed GSB-ViT. (a) The basic composition of the ViT model. 'L' indicates the number of transformer blocks. (b) The structure of a transformer block. 'MHSA' is the multi-head self-attention layer. (c) The structure of the GSB MHSA layer proposed in this paper. 'multiplied separately' is defined as Eq.~\ref{eq21}. (d) GSB matrix V and GSB attention matrix.}
\label{fig4}
\end{figure*}
To enhance the performance of the binarized attention matrix stated above, we propose a new binarization method named \textbf{G}roup \textbf{S}uperposition \textbf{B}inarization (GSB). \\
\indent The reason for skipping the translation operation of the attention matrix in the baseline is that the activation bias defined in XNOR-Net~\citep{rastegari2016xnor} is the average value of the activation. Adopting the same translation scheme as XNOR-Net will alter the distribution of values in the attention matrix from \textit{type-A} to \textit{type B} with zero mean (defined in~\ref{section3_1}), which is inconsistent with the non-negativity of the output of $softmax$ function. In addition, the fact that the activation value is always dependent on the input makes it extremely difficult to artificially set the most appropriate bias for the attention matrix. Considering that the scale factor and bias in XNOR-Net~\citep{rastegari2016xnor} cannot be well adapted to the binarization of attention matrix of ViT, we follow BiT~\citep{liu2022bit} to parameterize scale factor and bias, respectively. Furthermore, compared with the baseline, we can also improve efficiency by applying the learnable scale factor and bias to the activation because we can avoid the calculation of the mean and the norm of the matrix.
In our scheme, the bias is initialized with 0, and the most suitable bias can be obtained after model training and used to correct the distribution of values in $\mathbf{A}_{re}$ (reducing the error caused by the binarization operation of $\mathbf{Q}$ and $\mathbf{K}$, respectively).\\

\indent The architecture of our network is shown in Fig.~\ref{fig4}. For a real-valued attention matrix $\mathbf{A}_{re}\in \mathbb{R}^{H\times N\times N}$, $H$ and $N$ correspond to the largest head index and the largest image patch index of \textbf{M}ulti-\textbf{H}ead \textbf{S}elf \textbf{A}ttention (MHSA) mechanism, respectively. We first perform a translation operation on $\mathbf{A}_{re}$ as shown in Eq~\ref{eqoffset}. Thereafter, we define the corresponding GSB-binarization matrix of $\mathbf{A}_{ret}$ as Eq.~\ref{eq5}.
\begin{equation}
\label{eqoffset}
\centering
\begin{aligned}
\mathbf{A}_{ret}=\mathbf{A}_{re}-\boldsymbol{\Phi}, \\
\end{aligned}
\end{equation}
where $\boldsymbol{\Phi}\in \mathbb{R}^{H\times N\times N}$.
\begin{equation}
\label{eq5}
\centering
\begin{aligned}
\mathbf{A}_{bin}^{GSB}&=\alpha _0\mathbf{A}_{bin}+\sum_{i=1}^k{\alpha _i\mathbf{M}_i}, \\
\end{aligned}
\end{equation}
where $\mathbf{A}_{bin}\in \left\{ 0,1 \right\} ^{H\times N\times N}$ is defined in Eq~\ref{eq5-1}. $\alpha _0\in \mathbb{R}^{1}$ and $\alpha _i\in \mathbb{R}^{1}$ are learnable scaling factors.  $\mathbf{M}_{i}\in \left\{ 0,1 \right\} ^{H\times N\times N}$ is a mask indicating whether the value at a specific location of $\mathbf{A}_{ret}$ is larger than the corresponding threshold $\boldsymbol{\theta} _i$ (defined in Eq.~\ref{eq5-3}). The  $\mathbf{M}_{i}$ is defined in Eq.~\ref{eq5-2}.
\begin{equation}
\label{eq5-1}
\centering
\begin{aligned}
&\mathbf{A}_{bin}^{h,n,m}=clip\left( round\left( \frac{\mathbf{A}_{ret}^{h,n,m}}{\alpha _0} \right) ,0,1 \right),
\end{aligned}
\end{equation}
where $\mathbf{A}_{bin}^{h,n,m}$, $\mathbf{A}_{ret}^{h,n,m}$ are the element of $\mathbf{A}_{bin}$, $\mathbf{A}_{ret}$, respectively.   $h\in[1,H]$, $n\in[1,N]$, and $m\in[1,N]$.
\begin{equation}
\label{eq5-2}
\begin{aligned}
&\mathbf{M}_{i}^{h,n,m}=\left\{ \begin{matrix}
	1&		\mathbf{A}_{ret}^{h,n,m} >\boldsymbol{\Theta} _{i}^{h,n,m}\\
	0&		otherwise\\
\end{matrix} \right.,
\end{aligned}
\end{equation}
where $\mathbf{M}_{i}^{h,n,m}$, $\mathbf{A}_{ret}^{h,n,m}$, and $\boldsymbol{\Theta} _{i}^{h,n,m}$ are the element of $\mathbf{M}_{i}$, $\mathbf{A}_{ret}$, and $\boldsymbol{\Theta}_{i}$, respectively.

$\boldsymbol{\Theta} _i$ is an adaptive parameter determined by the maximum value of $\mathbf{A}_{ret}$ and a constant coefficient (Eq.~\ref{eq5-3}). $\boldsymbol{\Theta} _i$ does not need to compute gradient in the back-propagation process. \\
\begin{equation}
\label{eq5-3}
\begin{aligned}
\left( \mathbf{A}_{ret}^{\max} \right) ^{h,n}=\max \left( \mathbf{A}_{ret}^{h,n,1},\mathbf{A}_{ret}^{h,n,2}\cdots ,\mathbf{A}_{ret}^{h,n,N} \right) ,
\\
\boldsymbol{\Theta }_{i}^{h,n,m}=c_i\times \left( \mathbf{A}_{ret}^{\max} \right) ^{h,n}\,\, \left( m=1,2\cdots N \right) ,
\end{aligned}
\end{equation}
where $\mathbf{A}_{ret}^{\max}\in \mathbb{R}^{H\times N}, \mathbf{A}_{ret} \in \mathbb{R}^{H\times N\times N}$ and $\boldsymbol{\Theta}_{i} \in \mathbb{R}^{H\times N\times N}$. $c_i$ is defined as $0.5+\frac{0.4\times i}{k}$ ($k$ defined in Eq.\ref{eq5}), with a purpose to equally divide the interval $[0.5, 0.9]$.\\
\begin{figure}[htbp]
\setlength{\abovecaptionskip}{5pt}
\setlength{\belowcaptionskip}{0pt}
\centering
\includegraphics[width=5.0in]{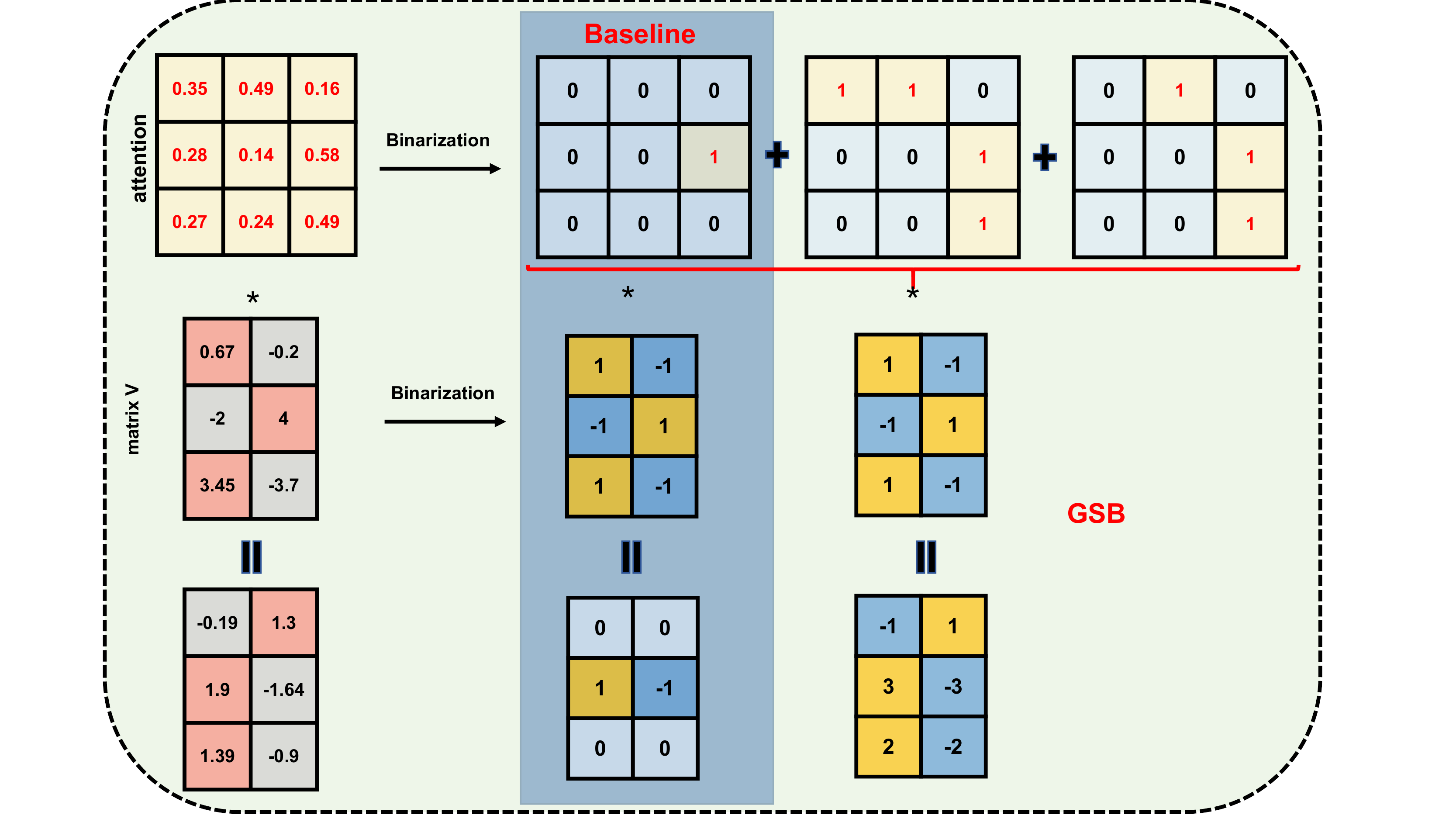}
\caption{Schematic diagram of the GSB process. For demonstration purposes, the scale factor is set to 1.}
\label{figgsbp}
\end{figure}
\indent Benefiting from the thresholded mask $\mathbf{M}_{i}$, the GSB method avoids the vanishing feature problem caused by binarizing the attention matrix. We take $k=2$ as an example and show it in Fig.~\ref{figgsbp}. Compared with the baseline, our GSB scheme can be a much better binary approximation of the value distribution of the real-valued attention matrix by the linear combination of the group of individually binarized attentions.\\

\indent Since the binarization function is not differentiable, we need to set a suitable gradient for it manually. For the proposed GSB method, directly applying the gradient of $\mathbf{A}_{bin}^{GSB}$ as the gradient of the $\mathbf{A}_{ret}$ can worsen the effect of gradient mismatch~\citep{lin2016overcoming,cai2017deep} (the difference between presumed and the actual gradient). Therefore, it is necessary to derive a suitable gradient calculation for the GSB process defined in Eq.\ref{eq5}. Likewise, we derive the gradient of $\alpha _0$ and $\alpha _i$, which are shown in Eq.~\ref{eq6} and Eq.~\ref{eq7}, respectively.\\
\begin{equation}
\label{eq6}
\begin{aligned}
&\frac{\partial \mathbf{L}}{\partial \alpha _0}=\frac{1}{H\times N\times N}\sum_{h,n,m}{\frac{\partial \mathbf{L}}{\partial \left( \mathbf{A}_{bin}^{GSB} \right) ^{h,n,m}}\frac{\partial \left( \mathbf{A}_{bin}^{GSB} \right) ^{h,n,m}}{\partial \alpha _0}},
\\
&\frac{\partial \left( \mathbf{A}_{bin}^{GSB} \right) ^{h,n,m}}{\partial \alpha _0}=\mathbf{A}_{bin}^{h,n,m}+\alpha _0\times\frac{\partial clip\left( round\left( \frac{\mathbf{A}_{ret}^{h,n,m}}{\alpha _0} \right) ,0,1 \right)}{\partial \alpha _0},
\\
&\begin{array}{c}
	STE\\
	\approx\\
\end{array}\left\{ \begin{matrix}
	\mathbf{A}_{bin}^{h,n,m}-\frac{\mathbf{A}_{ret}^{h,n,m}}{\alpha _0}&		0<\frac{\mathbf{A}_{ret}^{h,n,m}}{\alpha _0}<1\\
	\mathbf{A}_{bin}^{h,n,m}&		otherwise\\
\end{matrix} \right.,
\end{aligned}
\end{equation}
\noindent where $\frac{\partial \mathbf{L}}{\partial \left( \mathbf{A}_{bin}^{GSB} \right) ^{h,n,m}}$ is the gradient of the loss function with respect to $\left( \mathbf{A}_{bin}^{GSB} \right) ^{h,n,m}$, which comes from the deeper layer. $STE$ ~\citep{hinton2012neural,bengio2013estimating} is the straight through estimator.\\
\begin{equation}
\label{eq7}
\begin{aligned}
\frac{\partial \mathbf{L}}{\partial \alpha _i}&=\frac{1}{H\times N\times N}\sum_{h,n,m}{\frac{\partial \mathbf{L}}{\partial \left( \mathbf{A}_{bin}^{GSB} \right) ^{h,n,m}}\frac{\partial \left( \mathbf{A}_{bin}^{GSB} \right) ^{h,n,m}}{\partial \alpha _i}},
\\
\frac{\partial \left( \mathbf{A}_{bin}^{GSB} \right) ^{h,n,m}}{\partial \alpha _i}&=\mathbf{M}_{i}^{h,n,m}+\alpha _i\times\frac{\partial sign\left( relu\left( \mathbf{A}_{ret}^{h,n,m} -\boldsymbol{\Theta} _{i}^{h,n,m} \right) \right)}{\partial \alpha _i},
\\
&\begin{array}{c}
	STE\\
	\approx\\
\end{array}\mathbf{M}_{i}^{h,n,m},
\end{aligned}
\end{equation}
\indent As shown in Eq.~\ref{eq9_2}, Eq.~\ref{eq9}, and Eq.~\ref{eq10}, we derive the gradient of the loss function with respect to the real-valued input before binarization.\\
\begin{equation}
\label{eq9_2}
\frac{\partial \mathbf{L}}{\partial \mathbf{A}_{ret}}=\frac{\partial \mathbf{L}}{\partial \mathbf{A}_{bin}^{GSB}}\cdot\left( \mathbf{A}_0+\sum_{i=1}^k{\mathbf{A}_i} \right),
\end{equation}
where $\mathbf{A}_0$ and $\mathbf{A}_i$ refer to the components of the derivative of $\mathbf{A}_{bin}^{GSB}$ with respect to $\mathbf{A}_{ret}$. $\mathbf{A}_0$ and $\mathbf{A}_i$ correspond to $\mathbf{A}_{bin}$ and $\mathbf{M}_{i}$, respectively.  The operator $\cdot$ is the element-wise multiplication. \\ \par
\begin{footnotesize}
\begin{equation}
\begin{aligned}
\label{eq9}
&\mathbf{A}_{0}^{h,n,m}=\frac{\partial \left( \mathbf{A}_{bin}^{GSB} \right) ^{h,n,m}}{\partial \left( \mathbf{A}_{bin}^{h,n,m}\times \alpha _0 \right)}\times \frac{\partial \left( \mathbf{A}_{bin}^{h,n,m}\times \alpha _0 \right)}{\partial \mathbf{A}_{bin}^{h,n,m}}\times \frac{\partial \mathbf{A}_{bin}^{h,n,m}}{\partial \mathbf{A}_{ret}^{h,n,m}}
\\
&=1\times \alpha _0\times \frac{\partial clip\left( round\left( \frac{\mathbf{A}_{ret}^{h,n,m}}{\alpha _0} \right) ,0,1 \right)}{\partial \left( \mathbf{A}_{ret}^{h,n,m} \right)}
\\
&=\alpha _0\times \frac{\partial clip\left( round\left( \frac{\mathbf{A}_{ret}^{h,n,m}}{\alpha _0} \right) ,0,1 \right)}{\partial \frac{\mathbf{A}_{ret}^{h,n,m}}{\alpha _0}}\times \frac{\partial \frac{\mathbf{A}_{ret}^{h,n,m}}{\alpha _0}}{\partial \mathbf{A}_{ret}^{h,n,m}}
\\
&\begin{array}{c}
	STE\\
	\approx\\
\end{array}\left\{ \begin{matrix}
	1&		0< \frac{\mathbf{A}_{ret}^{h,n,m}}{\alpha _0}  <1\\
	0&		otherwise\\
\end{matrix} \right.
\end{aligned}
\end{equation}
\end{footnotesize}\par

So we have $\mathbf{A}_0\approx \mathbf{1}_{\left\{ 0<\left( \frac{\mathbf{A}_{ret}}{\alpha _0} \right) <1 \right\}},$ with
\begin{equation}
\label{eq6-1}
\begin{aligned}
&\mathbf{1}_{\left\{ 0<\frac{\mathbf{A}_{ret}}{\alpha _0}<1 \right\}}^{h,n,m}=\left\{ \begin{matrix}
	1&		0<\frac{\mathbf{A}_{ret}^{h,n,m}}{\alpha _0}<1\\
	0&		otherwise\\
\end{matrix} \right.,
\end{aligned}
\end{equation}

\indent The gradient of $\mathbf{M}_{i}$ can be calculated based on Eq.~\ref{eq5-2-2},
 \begin{equation}
\label{eq5-2-2}
\begin{aligned}
\mathbf{M}_i&=bool\left( \left( \mathbf{A}_{ret}-\boldsymbol{\Theta }_i \right) \right) ,
\\
bool\left( x \right)& =\left\{ \begin{matrix}
	1&		x>0\\
	0&		otherwise\\
\end{matrix}, \right.
\\
\frac{\partial bool\left( x \right)}{x}&=\left\{ \begin{matrix}
	1&		0<x< 1\\
	0&		otherwise\\
\end{matrix}, \right.
\end{aligned}
\end{equation}
where $\frac{\partial bool\left( x \right)}{x}$ is derived according to the STE.
\begin{footnotesize}
\begin{equation}
\begin{aligned}
\label{eq10}
\mathbf{A}_{i}^{h,n,m}&=\frac{\partial \left( \mathbf{A}_{bin}^{GSB} \right) ^{h,n,m}}{\partial \left( \mathbf{M}_{i}^{h,n,m}\times \alpha _i \right)}\times \frac{\partial \left( \mathbf{M}_{i}^{h,n,m}\times \alpha _i \right)}{\partial \mathbf{M}_{i}^{h,n,m}}\times \frac{bool\left( \mathbf{A}_{ret}^{h,n,m}-\boldsymbol{\Theta }_{i}^{h,n,m} \right)}{\mathbf{A}_{ret}^{h,n,m}-\boldsymbol{\Theta }_{i}^{h,n,m}}
\\
&\begin{array}{c}
	STE\\
	\approx\\
\end{array}\left\{ \begin{matrix}
	\alpha _i&		0<\left( \mathbf{A}_{ret}^{h,n,m}-\boldsymbol{\Theta }_{i}^{h,n,m} \right) < 1\\
	0&		otherwise\\
\end{matrix} \right.
\end{aligned}
\end{equation}
\end{footnotesize}\par
So we have $\mathbf{A}_i \approx \alpha _i\cdot \mathbf{1}_{\left\{ 0<\left( \mathbf{A}_{ret}-\boldsymbol{\Theta }_i \right) <1 \right\}}$.

Generally, GSB is sensitive to the initial value of the scaling factor. We obtain the initial value of each scale parameter by minimizing the quadratic approximation error as shown in Eq.~\ref{eq11} (where $k=2$ as an example), $\mathbf{a}_{ret}\in \mathbb{R}^p$ (the initial value of $\beta$ set to 0), $\mathbf{a}_{bin}\in \left\{ 0,1 \right\} ^p$, $\mathbf{m}_{1}\in \left\{ 0,1 \right\} ^p$, and $\mathbf{m}_{2}\in \left\{ 0,1 \right\} ^p$ represent the vectorized  $\mathbf{A}_{ret}$, $\mathbf{A}_{bin}$, $\mathbf{M}_{1}$, and $\mathbf{M}_{2}$, respectively, with $p=H \times N \times N$.
\begin{equation}
\begin{aligned}
\label{eq11}
\mathcal{J} \!\!\left( \alpha _0,\!\alpha _1,\!\alpha _2 \right) \!&=\!\!\left\| \mathbf{a}_{ret}\!\!-\!\left( \alpha _0\mathbf{a}_{bin}\!\!+\!\alpha _1\mathbf{m}_1\!\!+\!\alpha _2\mathbf{m}_2 \right) \right\| ^2.
\\
\left( \alpha _{0}^{*},\alpha _{1}^{*},\alpha _{2}^{*} \right) &=\!\!\underset{\alpha _0,\alpha _1,\alpha _2\in R^+}{arg\min}\mathcal{J} \left( \alpha _0,\alpha _1,\alpha _2 \right)
\end{aligned}
\end{equation}

 Eq.\ref{eq11} represents the error of approximation of the attention matrix $\mathbf{A}_{ret}$ by the Group Superposition Binarization. Supposing that $\alpha _{0}^{*}$, $\alpha _{1}^{*}$, and $\alpha _{2}^{*}$ are the solutions of $\mathcal{J} \!\!\left( \alpha _0,\!\alpha _1,\!\alpha _2 \right) \!$, then we can obtain the optimal initial values by taking the first-order derivatives of $\mathcal{J} \!\!\left( \alpha _0,\!\alpha _1,\!\alpha _2 \right) \!$ with respect to $\alpha _{0}$, $\alpha _{1}$, and $\alpha _{2}$, respectively, as shown in Eq.~\ref{eq12}.\\
\begin{equation}
\begin{aligned}
\label{eq12}
\frac{\partial \mathcal{J} \left( \alpha _0,\alpha _1,\alpha _2 \right)}{\partial \alpha _0}&=2\alpha _0\mathbf{a}_{bin}^{T}\mathbf{a}_{bin}+2\alpha _1\mathbf{a}_{bin}^{T}\mathbf{m}_1+2\alpha _2\mathbf{a}_{bin}^{T}\mathbf{m}_2-2\mathbf{a}_{ret}^{T}\mathbf{a}_{bin},
\\
\frac{\partial \mathcal{J} \left( \alpha _0,\alpha _1,\alpha _2 \right)}{\partial \alpha _1}&=2\alpha _1\mathbf{m}_{1}^{T}\mathbf{m}_1+2\alpha _0\mathbf{a}_{bin}^{T}\mathbf{m}_1+2\alpha _2\mathbf{m}_{1}^{T}\mathbf{m}_2-2\mathbf{a}_{ret}^{T}\mathbf{m}_1,
\\
\frac{\partial \mathcal{J} \left( \alpha _0,\alpha _1,\alpha _2 \right)}{\partial \alpha _2}&=2\alpha _2\mathbf{m}_{2}^{T}\mathbf{m}_{2}+2\alpha _0\mathbf{a}_{bin}^{T}\mathbf{m}_{2}+2\alpha _1\mathbf{m}_{1}^{T}\mathbf{m}_{2}-2\mathbf{a}_{ret}^{T}\mathbf{m}_{2}.
\end{aligned}
\end{equation}
Setting Eq.~\ref{eq12} equal to 0, we can get
\begin{equation}
\label{eq13}
\begin{aligned}
&\mathbf{a}_{bin}^{T}\mathbf{a}_{bin}\alpha _{0}^{*}+\mathbf{a}_{bin}^{T}\mathbf{m}_1\alpha _{1}^{*}+\mathbf{a}_{bin}^{T}\mathbf{m}_2\alpha _{2}^{*}=\mathbf{a}_{ret}^{T}\mathbf{a}_{bin},
\\
&\mathbf{a}_{bin}^{T}\mathbf{m}_1\alpha _{0}^{*}+\mathbf{m}_{1}^{T}\mathbf{m}_1\alpha _{1}^{*}+\mathbf{m}_{1}^{T}\mathbf{m}_2\alpha _{2}^{*}=\mathbf{a}_{ret}^{T}\mathbf{m}_1,
\\
&\mathbf{a}_{bin}^{T}\mathbf{m}_2\alpha _{0}^{*}+\mathbf{m}_{2}^{T}\mathbf{m}_1\alpha _{1}^{*}+\mathbf{m}_{2}^{T}\mathbf{m}_2\alpha _{2}^{*}=\mathbf{a}_{ret}^{T}\mathbf{m}_2.
\end{aligned}
\end{equation}

\indent Considering the problem that the $\mathbf{A}_{bin}$ defined in Eq.~\ref{eq5-1} cannot be decoupled from $\alpha _0$ and the problem that the value of $\mathbf{A}_{ret}$ may all be smaller than the threshold (0.5) of $round$ function (an example is shown in Fig.~\ref{figbasebi}), we set the initial value of $\alpha_{0}$ as the mean value of $\mathbf{A}_{ret}$. Then, Eq.~\ref{eq13} can be simplified as follow:
\begin{equation}
\label{eq13_2}
\begin{aligned}
&\mathbf{a}_{bin}^{T}\mathbf{m}_1\alpha _{0}^{*}+\mathbf{m}_{1}^{T}\mathbf{m}_1\alpha _{1}^{*}+\mathbf{m}_{1}^{T}\mathbf{m}_2\alpha _{2}^{*}=\mathbf{a}_{ret}^{T}\mathbf{m}_1,
\\
&\mathbf{a}_{bin}^{T}\mathbf{m}_2\alpha _{0}^{*}+\mathbf{m}_{2}^{T}\mathbf{m}_1\alpha _{1}^{*}+\mathbf{m}_{2}^{T}\mathbf{m}_2\alpha _{2}^{*}=\mathbf{a}_{ret}^{T}\mathbf{m}_2.
\end{aligned}
\end{equation}

\indent The optimal initial value can be easily obtained by solving Eq.~\ref{eq13_2}. From the properties of $\mathbf{a}_{bin}^{T}\mathbf{a}_{bin}=n_{\left\{ \mathbf{a}_{ret}\geqslant \alpha _{0}^{*}/2 \right\}}$, $\mathbf{m}_{1}^{T}\mathbf{m}_1=n_{\left\{ \mathbf{a}_{ret}\geqslant \boldsymbol{\theta _1} \right\}}$, and $\mathbf{m}_{2}^{T}\mathbf{m}_2=n_{\left\{ \mathbf{a}_{ret}\geqslant \boldsymbol{\theta _2} \right\}}$, where the format $n_{\left\{\mathbf{x}\geqslant t \right\}}$ is defined as the number of elements of vector $\mathbf{x}$ whose values are no smaller than a threshold $t$. $\boldsymbol{\theta _1}$ and $\boldsymbol{\theta _2}$ are the corresponding vectors of the thresholding matrices $\boldsymbol{\Theta} _1$ and $\boldsymbol{\Theta} _2$, respectively. The optimal initial value is shown in Eq.~\ref{eq16}.\\
\begin{equation}
\label{eq16}
\begin{aligned}
&\alpha _{0}^{*}\approx \frac{\left\| \mathbf{a}_{ret} \right\| _{l1}}{n_{\left\{ \mathbf{a}_{ret} \right\}}},
\\
&\alpha _{1}^{*}=\frac{\left\| \mathbf{a}_{ret}^{T}\cdot \mathbf{1}_{\left\{ \mathbf{a}_{ret}\geqslant \boldsymbol{\theta _2} \right\}} \right\| _{l1}-\left\| \mathbf{a}_{ret}^{T}\cdot \mathbf{1}_{\left\{ \mathbf{a}_{ret}\geqslant \boldsymbol{\theta _1} \right\}} \right\| _{l1}}{n_{\left\{ \mathbf{a}_{ret}\geqslant \boldsymbol{\theta _2} \right\}}-n_{\left\{ \mathbf{a}_{ret}\geqslant \boldsymbol{\theta _1} \right\}}}-\alpha _{0}^{*},
\\
&\alpha _{2}^{*}=\frac{\left\| \mathbf{a}_{ret}^{T}\cdot \mathbf{1}_{\left\{ \mathbf{a}_{ret}\geqslant \boldsymbol{\theta _2} \right\}} \right\| _{l1}}{n_{\left\{ \mathbf{a}_{ret}\geqslant \boldsymbol{\theta _2} \right\}}}-\left( \alpha _{0}^{*}+\alpha _{1}^{*} \right) .
\end{aligned}
\end{equation}
\subsection{Binarizing Matrix $\mathbf{V}$ based on GSB}
In order to improve the performance of the whole binary attention module, the part about matrix $\mathbf{V}$ is also important. The $sign$ function (Matrix V belongs to the \textit{type B} activation.) will convert numbers with different modulus values to the same one, which destroys the information contained in the features. We try to close the gap between the binarized matrix $\mathbf{V}$ and the full-precision one by GSB. Similar to the GSB of the attention matrix part, we binarize different parts of the matrix $\mathbf{V}$ separately by different thresholds. Thereafter, we superimpose the binarization results of different parts to enhance the representation ability of the binarization matrix $\mathbf{V}$. The forward GSB representation of $\mathbf{V}$ is as follows,
\begin{equation}
\begin{aligned}
\label{eq17}
\mathbf{V}_{bin}^{GSB}=\sum_{i=0}^k{\beta _i\mathbf{V}_{bin}^{i}}.
\\
\end{aligned}
\end{equation}
As shown in Eq.~\ref{eq17-1}, $\mathbf{V}_{bin}^{i}$ are components of $\mathbf{V}_{bin}^{GSB}$ and $\beta _i$ are learnable scale factors.
\begin{equation}
\begin{aligned}
\label{eq17-1}
&\mathbf{V}_{re}^{0}=\mathbf{V}_{re}-\boldsymbol{\Omega} ,\mathbf{V}_{bin}^{0}=sign\left( \mathbf{V}_{re}^{0} \right),
\\
&\mathbf{M}_i=\mathbf{1}_{\left\{ \left( \mathbf{V}_{re}^{0}>c_i\cdot v_{re}^{\max} \right) \cup \left( \mathbf{V}_{re}^{0}<c_i\cdot v_{re}^{\min} \right) \right\}},
\\
&\mathbf{V}_{re}^{i}=\mathbf{V}_{re}^{0}\cdot \mathbf{M}_{i},
\\
&\mathbf{V}_{bin}^{i}=\mathbf{V}_{bin}^{0}\cdot \mathbf{M}_{i},
\end{aligned}
\end{equation}
where $\mathbf{V}_{re}\in \mathbb{R} ^{H\times N\times C}$ is the corresponding real-valued matrix. $\boldsymbol{\Omega}  \in \mathbb{R} ^{H\times 1\times C}$ is a channel-wise learnable bias (one value per channel). During subtraction operation between $\mathbf{V}_{re}$ and $\boldsymbol{\Omega}$, the size of $\boldsymbol{\Omega}$ will be expanded as the size of $\mathbf{V}_{re}$ by repeating. $\mathbf{V}_{re}^{0}\in \mathbb{R} ^{H\times N\times C}$ is the translated product of $\mathbf{V}_{re}$. $c_i$ is the threshold coefficient defined as $0.5+\frac{0.4\times i}{k}$, which is used to equally split the interval $[0.5, 0.9]$ . $v_{re}^{\max}$ and $v_{re}^{\min}$ are two scalars referring to the maximum and minimum value of $\mathbf{V}_{re}^{0}$, respectively. Each $\mathbf{M}_i\in \left\{ 0,1 \right\} ^{H\times N\times C}$ is a threshold mask. In particular, $\mathbf{M}_0\in \left\{1 \right\} ^{H\times N\times C}$ is an all-one matrx.

\indent There is a problem that needs to be further considered. Because $\mathbf{V}_{bin}^{i}\,\, (i>0)$ contains a large number of 0, we cannot directly apply the $sign$ function to binarize it (see the definition of binarization process for \textit{type B}  activation in~\ref{section3_1}). Fortunately, the 0 values here are just a state of the mask to express nothing in the corresponding position of the matrix, and we can avoid the binarization of $\mathbf{V}_{bin}^{i}$ in both training and inference stages.

\begin{figure*}[htbp]
\setlength{\abovecaptionskip}{5pt}
\setlength{\belowcaptionskip}{0pt}
\centering
\includegraphics[width=6.5in]{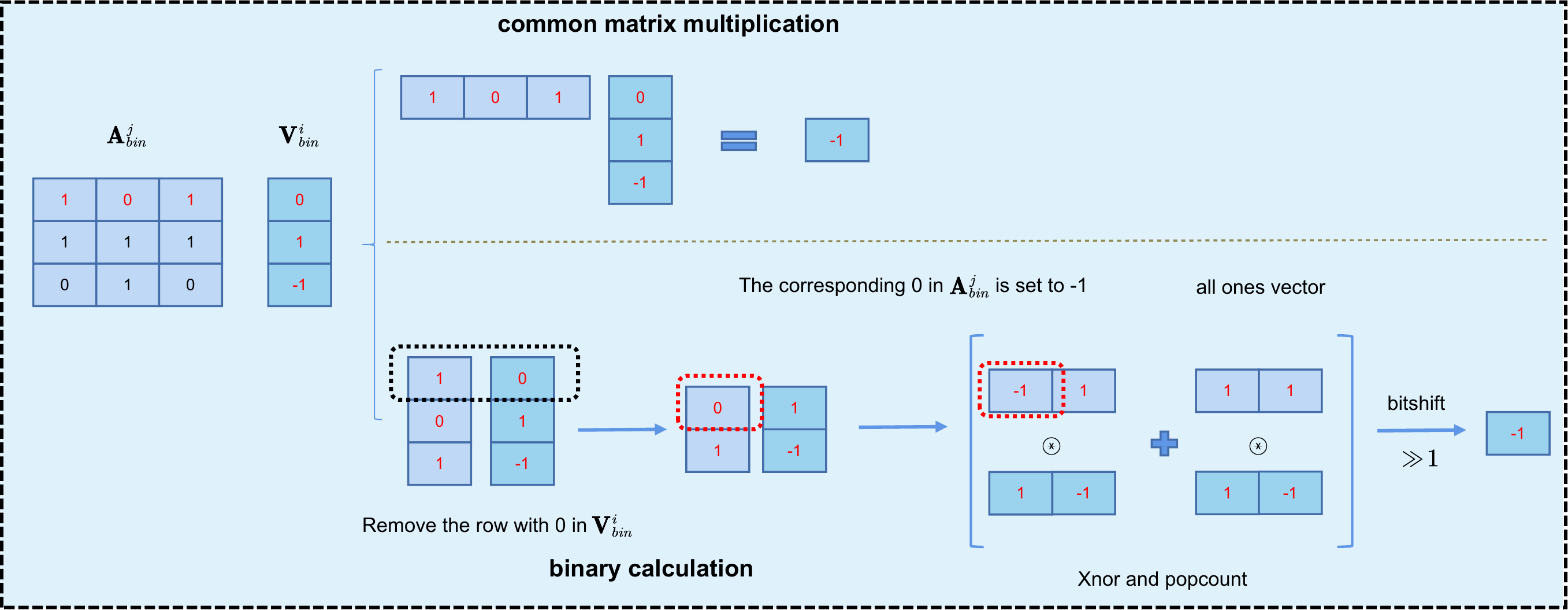}
\caption{The schematic diagram of the calculation method of the binarization matrix multiplication ($\mathbf{V}_{bin}^{i}$ and $\mathbf{A}_{bin}^{i}$ with 0-value, respectively. ) based on the $xnor$ and $popcount$ operations and the bitwise operation (proposed in Bibert~\citep{qin2022bibert}).}
\label{figoper}
\end{figure*}
During training,  thanks to the structure of GSB, this problem does not affect the training process of GSB because the common matrix multiplication is still applied during the training process, and the superimposed GSB matrix can erase the influence of the 0 value in $\mathbf{V}_{bin}^{i}$. During inference, to solve this problem when deploying this method to the edge-computing device, we use a scheme based on the bitwise operation proposed in BiBERT~\citep{qin2022bibert}. As shown in Fig~\ref{figoper}, for the 0 in the  $\mathbf{V}_{bin}^{i}$, we avoid binarizing it by skipping the $xnor$ and $popcount$ operation at the corresponding positions, and this can also reduce the number of operations.

Next, we need to re-derive the gradient calculation with respect to all scale factors. Apparently, it is not necessary to divide the full-precision value by the scaling factor for the forward process of the sign function (considering the property of the sign function,  which is often utilized in previous binarization work~\citep{liu2022bit,rastegari2016xnor,qin2022bibert}). However, since the scale factor affects the propagation range of the gradient in the back-propagation process, different from BiT~\citep{liu2022bit}, we re-derive the gradient calculation equation with respect to all scale factors, which are shown in Eq.~\ref{eq18} and Eq.~\ref{eq19}.\\
\begin{equation}
\label{eq18}
\begin{aligned}
\frac{\partial \mathbf{L}}{\partial \beta _i}=\frac{1}{H\times N \times C}\sum_{h,n,c}{\frac{\partial \mathbf{L}}{\partial \left( \mathbf{V}_{bin}^{GSB} \right) ^{h,n,c}}\frac{\partial \left( \mathbf{V}_{bin}^{GSB} \right) ^{h,n,c}}{\partial \beta _i}},
\end{aligned}
\end{equation}
where $\frac{\partial \left( \mathbf{V}_{bin}^{GSB} \right) ^{h,n,c}}{\partial \beta _i}$ is defined as Eq.~\ref{eq19}. $\left( \mathbf{V}_{bin}^{GSB} \right) ^{h,n,c}$ is the element of $\mathbf{V}_{bin}^{GSB}$. $h\in [1,H] ,n\in [1,N]$, and $c\in [1,C]$. Since the mask will bring a large number of 0 values, it is necessary to remove the gradient of the corresponding position. Therefore, $\frac{\partial \left( \mathbf{V}_{bin}^{GSB} \right) ^{h,n,c}}{\partial \beta _i}$ (where $i>0$) needs to be revised by the mask $\mathbf{M}_{i}$ with element-wise multiplication.\\ \par
\begin{footnotesize}
\begin{equation}
\label{eq19}
\begin{aligned}
&\frac{\partial \left( \mathbf{V}_{bin}^{GSB} \right) ^{h,n,c}}{\partial \beta _i}=\frac{\partial \left( sign\left( \frac{\left( \mathbf{V}_{re}^{i} \right) ^{h,n,c}}{\beta _i} \right) \cdot \beta _i \right)}{\partial \beta _i}
\\
&=sign\left( \frac{\left( \mathbf{V}_{re}^{i} \right) ^{h,n,c}}{\beta _i} \right) +\beta _i\cdot \frac{\partial sign\left( \frac{\left( \mathbf{V}_{re}^{i} \right) ^{h,n,c}}{\beta _i} \right)}{\partial \beta _i}
\\
&\begin{array}{c}
	STE\\
	\approx\\
\end{array}\left\{ \begin{matrix}
	sign\left( \left( \mathbf{V}_{re}^{i} \right) ^{h,n,c} \right) -\frac{\left( \mathbf{V}_{re}^{i} \right) ^{h,n,c}}{\beta _i}&		-1<\frac{\left( \mathbf{V}_{re}^{i} \right) ^{h,n,c}}{\beta _i}<1\\
	sign\left( \left( \mathbf{V}_{re}^{i} \right) ^{h,n,c} \right)&		otherwise\\
\end{matrix} \right..
\end{aligned}
\end{equation}
\end{footnotesize}
\indent We derive the gradient of the loss function with respect to the $\mathbf{V}_{re}^{0}$ before binarization.\\
\begin{equation}
\label{eq20-1}
\begin{aligned}
\frac{\partial \mathbf{L}}{\partial \mathbf{V}_{re}^{0}}=\frac{\partial \mathbf{L}}{\partial \mathbf{V}_{bin}^{GSB}}\cdot \sum_{i=0}^k{\mathbf{V}_i},
\end{aligned}
\end{equation}
where $\mathbf{V}_i$ is the gradient components corresponding to the components of $\mathbf{V}_{bin}^{GSB}$, which is shown in Eq.~\ref{eq20-2} and Eq.~\ref{eq20-3}.\\
\begin{equation}
\label{eq20-2}
\begin{aligned}
\mathbf{V}_{0}^{h,n,c}&=\frac{\partial \left( \mathbf{V}_{bin}^{GSB} \right) ^{h,n,c}}{\partial \left( \mathbf{V}_{bin}^{0} \right) ^{h,n,c}}\times \frac{\partial sign\left( \frac{\left( \mathbf{V}_{re}^{0} \right) ^{h,n,c}}{\beta _0} \right)}{\partial \left( \mathbf{V}_{re}^{0} \right) ^{h,n,c}}
\\
&=\beta _0\times \frac{\partial sign\left( \frac{\left( \mathbf{V}_{re}^{0} \right) ^{h,n,c}}{\beta _0} \right)}{\partial \frac{\left( \mathbf{V}_{re}^{0} \right) ^{h,n,c}}{\beta _0}}\times \frac{1}{\beta _0}
\\
&\begin{array}{c}
	STE\\
	\approx\\
\end{array}\left\{ \begin{matrix}
	1&		-1<\frac{\left( \mathbf{V}_{re}^{0} \right) ^{h,n,c}}{\beta _0}<1\\
	0&		otherwise\\
\end{matrix} \right. ,
\\
&\text{So we have}~~\mathbf{V}_0=\mathbf{1}_{\left\{ -1<\frac{\mathbf{V}_{re}^{0}}{\beta _0}<1 \right\}},
\end{aligned}
\end{equation}
where $\mathbf{V}_{0}^{h,n,c}$ is the element of $\mathbf{V}_{0}$. $\mathbf{V}_{i} (i>0)$ is defined as Eq.~\ref{eq20-3}.\\
\begin{equation}
\label{eq20-3}
\begin{aligned}
&\mathbf{V}_{i}^{h,n,c}=\frac{\partial \left( \mathbf{V}_{bin}^{GSB} \right) ^{h,n,c}}{\partial \left( \mathbf{V}_{bin}^{i} \right) ^{h,n,c}}\times \frac{\partial sign\left( \frac{\left( \mathbf{V}_{re}^{0} \right) ^{h,n,c}}{\beta _i} \right)}{\partial \left( \mathbf{V}_{re}^{0} \right) ^{h,n,c}}\times \mathbf{M}_{i}^{h,n,c}
\\
&=\beta _i\times \frac{\partial sign\left( \frac{\left( \mathbf{V}_{re}^{0} \right) ^{h,n,c}}{\beta _i} \right)}{\partial \frac{\left( \mathbf{V}_{re}^{0} \right) ^{h,n,c}}{\beta _i}}\times \frac{1}{\beta _i}\times \mathbf{M}_{i}^{h,n,c}
\\
&\begin{array}{c}
	STE\\
	\approx\\
\end{array}\left\{ \begin{matrix}
	1&		\left( \frac{\left( \mathbf{V}_{re}^{0} \right) ^{h,n,c}}{\beta _i}\in \left[ -1,1 \right] \right) \cap \left( \mathbf{M}_{i}^{h,n,c}\ne 0 \right)\\
	0&		otherwise\\
\end{matrix} \right. ,
\\
&\text{So we have}~~\mathbf{V}_i=\mathbf{1}_{\left\{ \left( \frac{\left( \mathbf{V}_{re}^{0} \right)}{\beta _i}\in \left[ -1,1 \right] \right) \cap \left( \mathbf{M}_i\ne 0 \right) \right\}}.
\end{aligned}
\end{equation}
\indent For the initial value of scale factors for the GSB matrix $\mathbf{V}$, we apply the same method as Eq.\ref{eq11} to obtain the linear equations shown in Eq.\ref{eq11-22}.\\\par
\begin{footnotesize}
\begin{equation}
\label{eq11-22}
\begin{aligned}
\left( \mathbf{v}_{bin}^{0} \right) ^T\mathbf{v}_{bin}^{0}\beta _{0}^{*}+\left( \mathbf{v}_{bin}^{0} \right) ^T\mathbf{v}_{bin}^{1}\beta _{1}^{*}+\left( \mathbf{v}_{bin}^{0} \right) ^T\mathbf{v}_{bin}^{2}\beta _{2}^{*}=\mathbf{v}_{re}^{T}\mathbf{v}_{bin}^{0},
\\
\left( \mathbf{v}_{bin}^{0} \right) ^T\mathbf{v}_{bin}^{1}\beta _{0}^{*}+\left( \mathbf{v}_{bin}^{1} \right) ^T\mathbf{v}_{bin}^{1}\beta _{1}^{*}+\left( \mathbf{v}_{bin}^{1} \right) ^T\mathbf{v}_{bin}^{2}\beta _{2}^{*}=\mathbf{v}_{re}^{T}\mathbf{v}_{bin}^{1},
\\
\left( \mathbf{v}_{bin}^{0} \right) ^T\mathbf{v}_{bin}^{2}\beta _{0}^{*}+\left( \mathbf{v}_{bin}^{2} \right) ^T\mathbf{v}_{bin}^{1}\beta _{1}^{*}+\left( \mathbf{v}_{bin}^{2} \right) ^T\mathbf{v}_{bin}^{2}\beta _{2}^{*}=\mathbf{v}_{re}^{T}\mathbf{v}_{bin}^{2}.
\end{aligned}
\end{equation}
\end{footnotesize}
\indent We can also optimize the solution of the above formula according to the value of $\left( \mathbf{v}_{bin}^{i} \right) ^T\mathbf{v}_{bin}^{j}$ and $\mathbf{v}_{re}^{T}\mathbf{v}_{bin}^{i}$. To derive the above two values, we first need to reshape each mask $\mathbf{M}_{i}$ to the $\mathbf{m}_i$, which has the same dimension as $\mathbf{v}_{bin}^{i}$. In particular, $\mathbf{m}_{0}$ is an all-one vector. Then, we can conclude that $\left( \mathbf{v}_{bin}^{i} \right) ^T\mathbf{v}_{bin}^{j}=n\left( \mathbf{v}_{re}\cdot \mathbf{m}_{\max \left( i,j \right)} >0\right) $ and $\mathbf{v}_{re}^{T}\mathbf{v}_{bin}^{i}=\left\| \mathbf{v}_{re}\cdot \mathbf{m}_i \right\| _{l1}$, where the format $n_{\left\{\mathbf{x}> t \right\}}$ is defined as the number of elements of vector $\mathbf{x}$ whose values are larger than a threshold $t$. The value of $max(i,j)$ is the larger one of $i$ and $j$. Therefore,
if $i>j$, then $max(i,j)=i$, $\mathbf{m}_{\max \left( i,j \right)}=\mathbf{m}_{i}$. Otherwise, $\mathbf{m}_{\max \left( i,j \right)}=\mathbf{m}_{j}$.  As shown in Eq\ref{eq16-V}, according to the above analysis, we can get the optimal estimate of $\beta _{0}^{*}$, $\beta _{1}^{*}$, and $\beta _{2}^{*}$.
\begin{equation}
\label{eq16-V}
\begin{aligned}
&\beta _{0}^{*}=\frac{\left\| \mathbf{v}_{re}^{T}\cdot m_0 \right\| _{l1}-\left\| \mathbf{v}_{re}^{T}\cdot m_1 \right\| _{l1}}{n_{\left\{ \mathbf{v}_{re}^{T}\cdot m_0 >0 \right\}}-n_{\left\{ \mathbf{v}_{re}^{T}\cdot m_1 >0\right\}}},
\\
&\beta _{1}^{*}=\frac{\left\| \mathbf{v}_{re}^{T}\cdot m_1 \right\| _{l1}-\left\| \mathbf{v}_{re}^{T}\cdot m_2 \right\| _{l1}}{n_{\left\{ \mathbf{v}_{re}^{T}\cdot m_1 >0\right\}}-n_{\left\{ \mathbf{v}_{re}^{T}\cdot m_2 >0\right\}}}-\alpha _{0}^{*},
\\
&\beta _{2}^{*}=\frac{\left\| \mathbf{v}_{re}^{T}\cdot m_2 \right\| _{l1}}{n_{\left\{ \mathbf{v}_{re}^{T}\cdot m_2 >0\right\}}}-\left( \alpha _{0}^{*}+\alpha _{1}^{*} \right) .
\end{aligned}
\end{equation}
\indent With the above settings, the output of the GBS-MHSA attention module with the GSB algorithm can be calculated as:\par
\begin{equation}
\label{eq21}
\begin{aligned}
\mathbf{Y}&=\mathbf{A}_{bin}^{GSB}*\mathbf{V}_{bin}^{GSB}
\\
&=\sum_{i=0}^k{\sum_{j=0}^k{\alpha _i\beta _j\left( \mathbf{A}_{bin}^{i}\boxtimes \mathbf{V}_{bin}^{j} \right)}},
\end{aligned}
\end{equation}
where $\mathbf{A}_{bin}^{i}$ and $\mathbf{V}_{bin}^{j}$ refer to the components of $\mathbf{A}_{bin}^{GSB}$ and $\mathbf{V}_{bin}^{GSB}$, respectively. $\alpha_i$ and $\beta_j$ are the scale factors of the corresponding $\mathbf{A}_{bin}^{i}$ and $\mathbf{V}_{bin}^{j}$, respectively. $\mathbf{A}_{bin}^{i} \left( i\in \left( 1,2\cdots k \right) \right) $ indicates mask $\mathbf{M}_i$ applied in the GSB attention matrix. The operator $\boxtimes $ means the matrix multiplication replaced by the operation defined in Fig~\ref{figoper}.\\
\indent Almost all parameters in ViT exist in the linear layer. The binarization process of weight in the linear layer is consistent with our baseline. For the activation of the linear layer, we apply learnable bias and learnable scale factor to complete the translation and scale operation of the activation, respectively.

\subsection{Distillation for training on small datasets from scratch}
Compared with CNN, owing to a lack of inductive bias, the transformer model requires a huge amount of data to obtain a relatively good performance~\citep{touvron2021training}. Due to the insufficient amount of data, the full-precision transformer often suffers from the overfitting problem on the small dataset. Although the binarized model can relieve the overfitting problem to a certain extent, its weak representation ability of the binary weights and the introduced binarization error still significantly degrade the model's performance. Meanwhile, the gradient error of the binarized model during training also makes it difficult to converge. In order to solve the above problems, we use distillation learning to optimize the training process of the binarization model. We select the corresponding full-precision transformer pre-trained on Imagenet-1K as the teacher model (fine-tuned to each dataset). The loss function is shown in Eq.~\ref{eq22}.
\begin{equation}
\label{eq22}
L=\left( 1-\lambda  \right) L_{cls}+\lambda L_{dis},
\end{equation}
where $L_{cls}$ means the cross-entropy loss between the output of the model's class token and the label. $L_{dis}$ is the cross entropy loss between the output of the teacher network and the student network, respectively. We use the hard-label distillation mechanism of DeiT, where the output of the teacher network plays the same role as the labels. Therefore, we set $\lambda$ to 0.5 to balance the sizes of the two loss components. This distillation mechanism is parameter-free and conceptually simpler. Meanwhile, compared to using soft-label distillation, our method achieves better performance (71.10\% VS 69.40\%) on the CIFAR-100 dataset by employing the hard-label distillation mechanism.

\indent Similar to most training strategies for binarized models~\citep{liu2022bit,liu2020reactnet}, we apply the Two-Stage Training Strategy (TSTS) for the training process of GSB-ViT. In the first stage, we only binarize the weights of the model. In contrast, we binarize all the weights and the activations in the second stage. After the first stage of training is over, we use the trained weights as the initial values of the model‘s weights in the second stage.

\begin{figure}[htbp]
\setlength{\abovecaptionskip}{5pt}
\setlength{\belowcaptionskip}{0pt}
\centering
\includegraphics[width=3.0in]{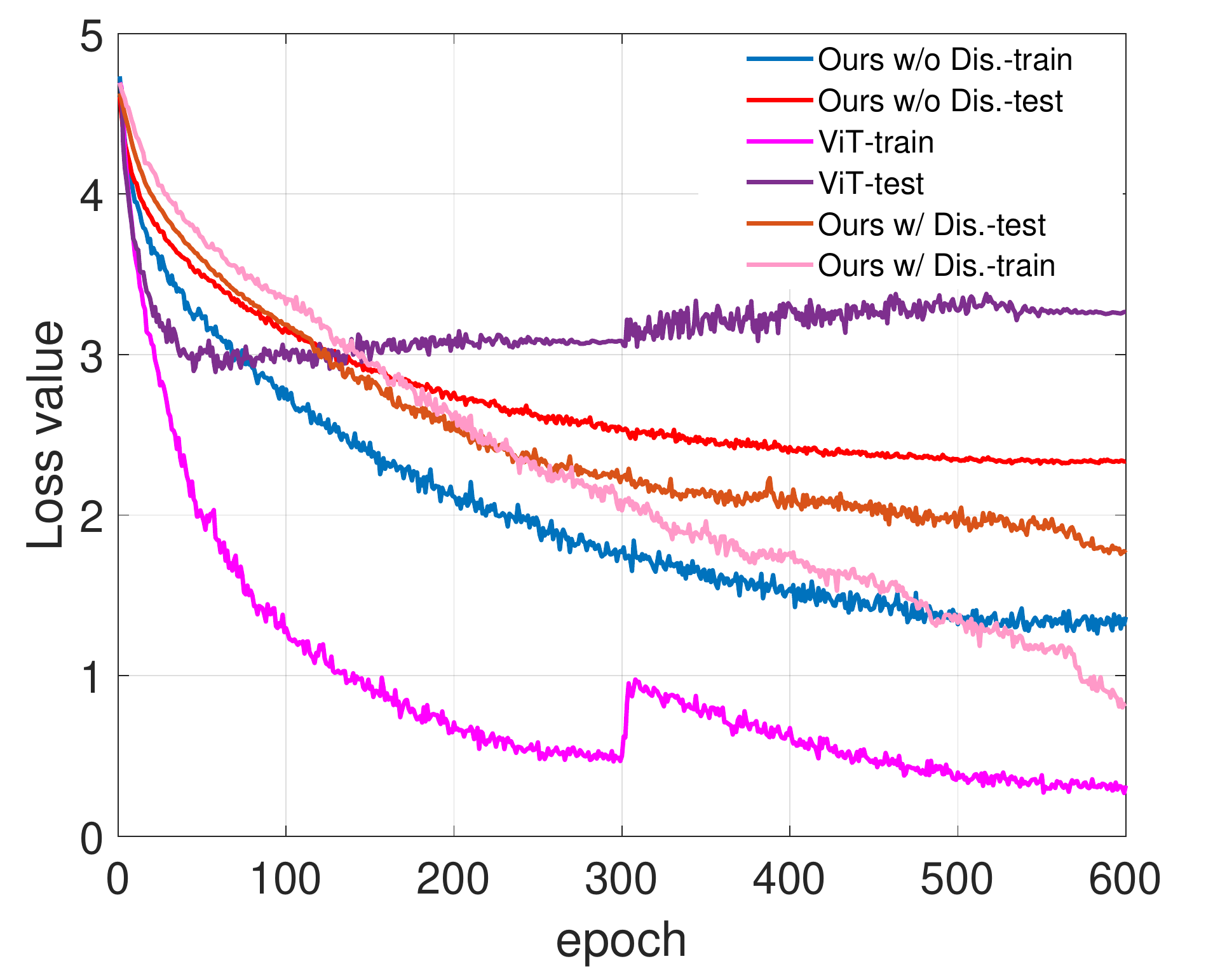}
\caption{The loss curve of the full-precision ViT model (The DeiT-small without distillation can be seen as the ViT model with strong data augmentation), our proposed method w/o distillation, and our proposed method with distillation trained from scratch on the Oxford-Flowers102 dataset, respectively. To ensure the fairness of the comparison, we only apply one training stage to all models in this comparative experiment.}
\label{fig5}
\end{figure}

\indent During the model training process, we find that the full-precision ViT model is prone to be affected by overfitting problems on small datasets. In contrast, our model can relieve the overfitting problem to a certain extent. As shown in Fig.~\ref{fig5}, we can find out that the training loss of full-precision ViT drop to a very low state, which can show that full-precision weights and activations in ViT are expressive enough to learn from the dataset. The weights and activations of the binary model are relatively weaker because of the existence of binarization error. However, the test loss of full-precision ViT is not satisfying. We adjust its learning rate after 300 epochs and still cannot change this phenomenon. In contrast,  the situation has improved significantly for the binary model.

\indent To explain this phenomenon, we can focus on the difference between the binarized model and the full-precision one. The binarization process inevitably introduces the binarization error, which is similar to the quantization error~\citep{jung2021quantization,fang2020post,li2022q} and usually expressed as the difference between the full-precision value and the corresponding binarization one. On the other hand, the binarization process constrains the solution space of the parameters and narrows the scope of the solution space (limiting the value of each parameter from the real number set to binary values). Therefore, the binarization process can be seen as a special kind of implicit regularization, preventing the model which is trained on small datasets from overfitting.  Distillation further makes the output of the student model and the teacher model close and thus reduces the binarization error. In summary, the binarization process plays a favorable role in the robustness of the binary model. Meanwhile, distillation loss will reduce the binarization error to improve the performance of the model continuously. Therefore, the binarization model is more suitable for the situation where the amount of data is insufficient.

\section{Experiment}\label{section5}
\subsection{Datasets and Implementation Details }
\noindent \textbf{Datasets:} We evaluate our method on three small datasets including Chaoyang ~\citep{zhu2021hard}, CIFAR-100~\citep{krizhevsky2009learning}, and Oxford-Flowers102 ~\citep{nilsback2008automated}. The  Chaoyang~\citep{zhu2021hard} dataset includes 4021 training images for 4 classes and 2139 test images. CIFAR-100~\citep{krizhevsky2009learning} includes 50,000 training images and 10,000 test images for 100 classes. Oxford-Flowers102~\citep{nilsback2008automated} contains 2040 training images and 6149 test images for 102 classes.

\noindent \textbf{Implementation Details:} We resize the image resolution to the size of $224 \times 224$ for the input of the model. For data augmentation, we follow the scheme proposed in DeiT~\citep{touvron2021training}. The Adam optimizer combined with the cosine annealing learning-rate decay and an initial learning rate of $5 \times 10^{-4}$ is utilized to train the GSB-Vision Transformer. For distillation learning, we adopt the multi-class token distillation scheme proposed by Touvron et al.~\citep{touvron2021training}. We apply 2 TITAN-RTX-48GB GPUs with a 128-batch size to complete the model training process for CIFAR-100 datasets~\citep{krizhevsky2009learning} and 1 TITAN-RTX-48GB GPU with 64-batch size for Oxford-Flowers102 ~\citep{nilsback2008automated} and  Chaoyang~\citep{zhu2021hard}, respectively. Compared with the corresponding full-precision network, the binary network requires more iterative training times~\citep{qin2023bibench}. Considering that the binary network is more difficult to converge than the full-precision model, we set 900 epochs during the training process without a warm-up technique for each dataset (600 epochs for the first stage and 300 epochs for the second stage).

\subsection{Results On CIFAR-100}

\begin{figure}[htbp]
\setlength{\abovecaptionskip}{5pt}
\setlength{\belowcaptionskip}{0pt}
\centering
\includegraphics[width=6.2in]{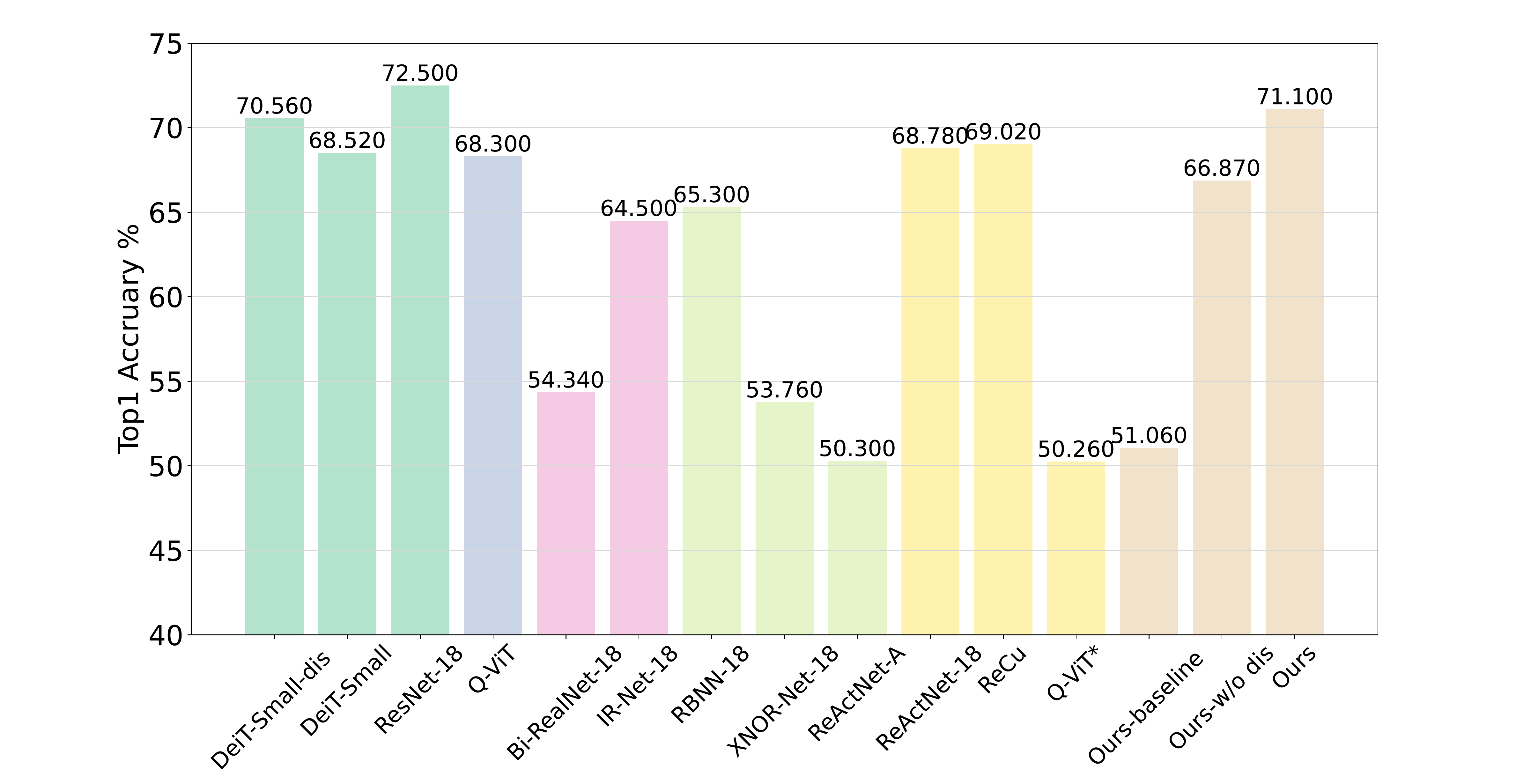}
\caption{Comparison results for classification on CIFAR-100.}
\label{figcifar100}
\end{figure}

As shown in Tab.~\ref{tab1} and Fig.~\ref{figcifar100}, columns W-A indicates the number of bit  of weights and activations. The method that achieves the best performance at each different level of bit is marked in bold. The column $Disti.$ means whether the model applies knowledge distillation during training. The column $Trans.$ refers to whether the architecture of the model is transformer or CNN. The column $OPs.$ indicates the number of operation of each method, which can evaluate the computational complexity of a model. Our baseline model has a large performance gap compared with the full-precision model. Combined with the GSB method, the performance gap is significantly narrowed, which confirmed the effectiveness of our proposed method. With the distillation scheme, our binarized model even outperforms the corresponding full-precision model with the guidance of the same teacher model.

The reason for this phenomenon is that the full-precision model suffers from overfitting in the later stage of training (after 500 epochs). Compared with the binarized CNN, our algorithm also achieves SOTA performance.
This experiment demonstrates the effectiveness of the proposed GSB-vision transformer model whose classification performance is better than the compared binary methods and even exceeds its full-precision counterpart.
\begin{table}[htbp]
    	\caption{Comparison results for classification on CIFAR-100. Q-ViT* means changing the bit level to 1 bit on the framework of Q-ViT and applying the $sign$ function to replace the quantization function used in Q-ViT.}
		\label{tab1}
		\centering
		\begin{tabular}{c|c|c|c|c|c|c}  
			\hline
			\hline
			Methods                                                    &Disti.       &Trans.   & W-A          &  OPs($\times 10^8$)  &  Top-1(\%)          &  Top-5(\%)       \\
			\hline
            \hline
			DeiT-Small~\citep{touvron2021training}                     &\ding{52}    &\ding{52}& 32-32          & 46.0     & 70.56              & 92.10       \\
			\hline
            DeiT-Small~\citep{touvron2021training}                     &\ding{55}    &\ding{52}& 32-32           & 46.0     & 68.52              & 91.35        \\
			\hline
   		ResNet-18~\citep{he2016deep}                               &\ding{55}    &\ding{55}& 32-32           & 18.1   & \textbf{72.50}     & 92.75         \\
			\hline
			Q-ViT~\citep{li2022q}                                      &\ding{52}    &\ding{52} & 2-2            & -     & \textbf{68.30}     & 91.04       \\
            \hline
			Bi-RealNet-18~\citep{liu2018bi}                            &\ding{52}    &\ding{55} & 1-1            & 1.63    & 54.34              & 82.24       \\
            \hline
            IR-Net-18~\citep{qin2020forward}                           &\ding{52}    &\ding{55} & 1-1            & -      & 64.50              & 85.43      \\
            \hline
            RBNN-18~\citep{lin2020rotated}                             &\ding{52}    &\ding{55} & 1-1            & -      & 65.30              & 87.65      \\
            \hline
			XNOR-Net-18~\citep{rastegari2016xnor}                      &\ding{55}    &\ding{55} & 1-1            & 1.67    & 53.76              & 81.58         \\
            \hline
			ReActNet-A~\citep{liu2020reactnet}                             &\ding{52}    &\ding{55} & 1-1           & 0.87    & 50.30              & 81.22         \\
            \hline
            ReActNet-18~\citep{liu2020reactnet}                            &\ding{52}    &\ding{55} & 1-1          & 1.63       & 68.78              & 91.28       \\
            \hline
            ReCu~\citep{xu2021recu}                                    &\ding{52}    &\ding{55} & 1-1           & -        & 69.02              & 91.89     \\
            \hline
			Q-ViT*~\citep{li2022q}                                     &\ding{52}    &\ding{52} & 1-1           & 1.41    & 50.26              & 80.06        \\
            \hline
			Ours-baseline                                              &\ding{55}    &\ding{52} & 1-1           & 1.41      & 51.06              & 80.78       \\
            \hline
			Ours-w/o disti                                             &\ding{55}    &\ding{52} & 1-1           & 1.68      & 66.87              & 89.43       \\
            \hline
			Ours                                                       &\ding{52}    &\ding{52} & 1-1           & 1.68       & \textbf{71.10}     & 92.45      \\
			\hline
			\hline
		\end{tabular}
	\end{table}
\subsection{Results On Oxford-Flowers102}

\begin{figure}[htbp]
\setlength{\abovecaptionskip}{5pt}
\setlength{\belowcaptionskip}{0pt}
\centering
\includegraphics[width=5.0in]{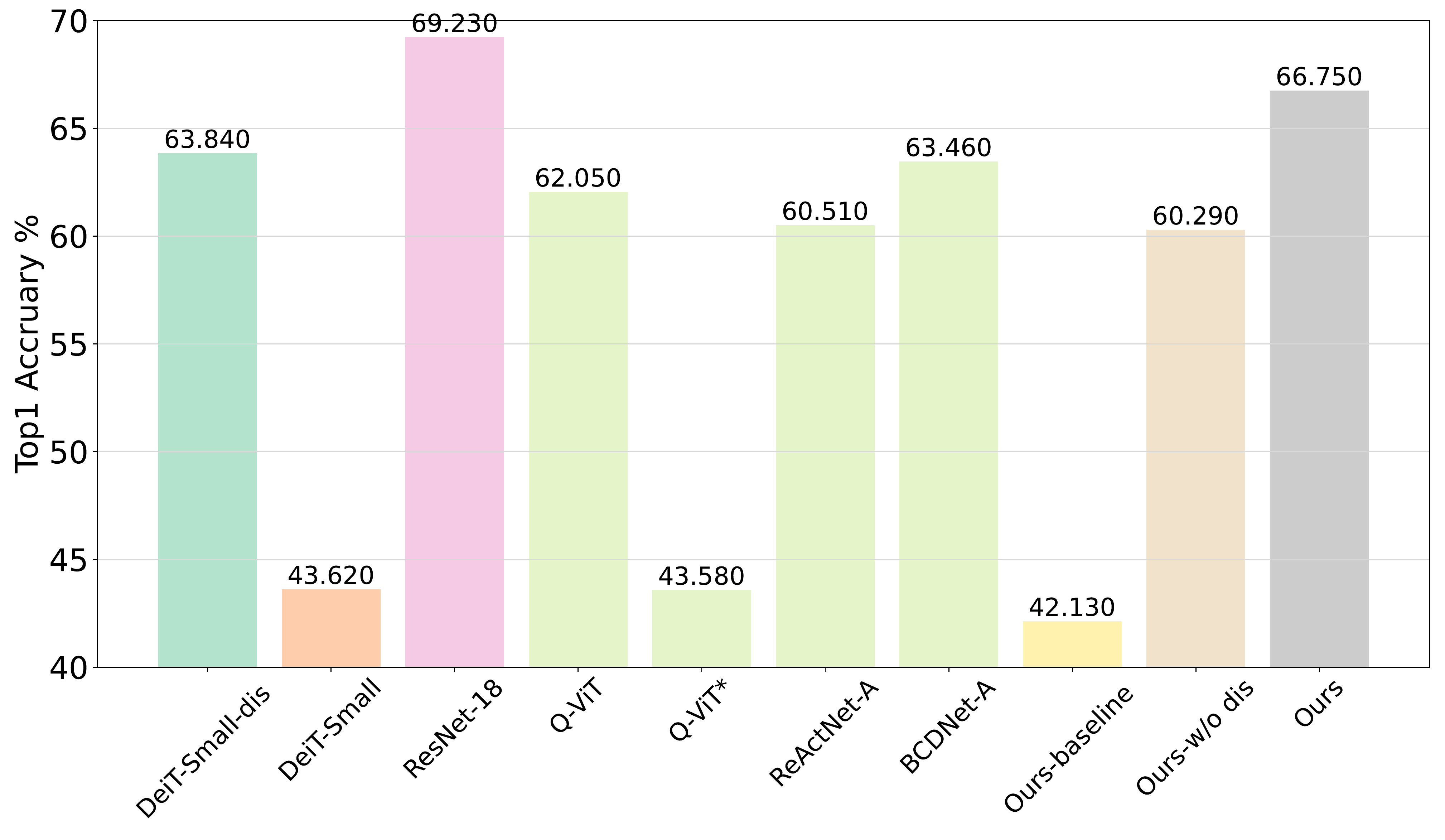}
\caption{Comparison results for classification on Oxford-Flowers102.}
\label{figflower}
\end{figure}

The split of the training and test set of the Oxford-Flowers102 is not balanced compared to that in the Chaoyang ~\citep{zhu2021hard} and CIFAR-100~\citep{krizhevsky2009learning} datasets. Only 2040 training images are not enough and will cause serious overfitting of ViT. Note that the DeiT model without distillation can be seen as a ViT model with special data augmentation. Compared with the full-precision CNN, ViT is difficult to be trained well with 2040 training images. Although the training loss was minimized, the overfitting problem results in the little robustness of ViT. The experimental results are shown in Tab.~\ref{tab2} and Fig.~\ref{figflower}. Influenced by the excessive binarization error, the training loss of our baseline model cannot be reduced to a low level. On the contrary, our GSB method without knowledge distillation can mitigate the effects of overfitting (60.29\% vs 43.62\%) and reduce the performance degradation caused by the binarization error (60.29\% vs 42.13\%). However, the result is still unsatisfactory. Coupled with the distillation method, the performance of our GSB model far exceeds its corresponding full-precision network. Compare with the SOTA BNN methods (ReActNet-A and BCDNet-A), our approach still has performance advantages.  Consistent with the other experiments, there is still some performance gap between our GSB model and the full-precision CNN model. This experiment validates that the binarization process has a certain positive effect on small data training tasks, which also reveals that our method is more suitable for the case where the amount of data is severely insufficient.
\begin{table}[htbp]
    	\caption{Comparison results for classification on Oxford-Flowers102.}
		\label{tab2}
		\centering

		\begin{tabular}{c|c|c|c|c|c|c}  
			\hline
			\hline
			Methods                                  &Disti.    &Trans.& W-A   & Params number    &  Top-1(\%)     &  Top-5(\%)     \\
			\hline
                \hline
			DeiT-Small~\citep{touvron2021training}                     &\ding{52}    &\ding{52}& 32-32   &21.8M    & 63.84 & 84.21  \\
			\hline
			DeiT-Small~\citep{touvron2021training}                      &\ding{55}    &\ding{52}& 32-32  &21.8M     & 43.62  & 69.39 \\
			\hline
			ResNet-18~\citep{he2016deep}                               &\ding{55}    &\ding{55} &32-32   &11.3M     & \textbf{69.23}  & 90.05  \\
            \hline
            Q-ViT~\citep{li2022q}                                     &\ding{52}    &\ding{52} & 2-2     &21.8M   & 62.05   & 82.64 \\
            \hline
            Q-ViT*~\citep{li2022q}                                     &\ding{52}    &\ding{52} & 1-1    &21.8M    & 43.58  & 69.08  \\
            \hline
            ReActNet-A~\citep{liu2020reactnet}                                    &\ding{52}    &\ding{55} & 1-1     &28.4M   & 60.51 & 80.98   \\
            \hline
            BCDNet-A~\citep{xing2022towards}                                     &\ding{52}    &\ding{55} & 1-1    &44.3M    & 63.46 & 83.20   \\
            \hline
			Ours-baseline                                             &\ding{55}    &\ding{52} & 1-1     &21.8M   & 42.13  & 68.05 \\
            \hline
			Ours-w/o disti                                            &\ding{55}    &\ding{52} & 1-1     &24.5M   & 60.29  & 79.54 \\
            \hline
			Ours                                                      &\ding{52}    &\ding{52} & 1-1     &24.5M   & \textbf{66.75} & 85.33  \\
			\hline
			\hline
		\end{tabular}
	\end{table}
\subsection{Results On Chaoyang}

\begin{figure}[htbp]
\setlength{\abovecaptionskip}{5pt}
\setlength{\belowcaptionskip}{0pt}
\centering
\includegraphics[width=5.0in]{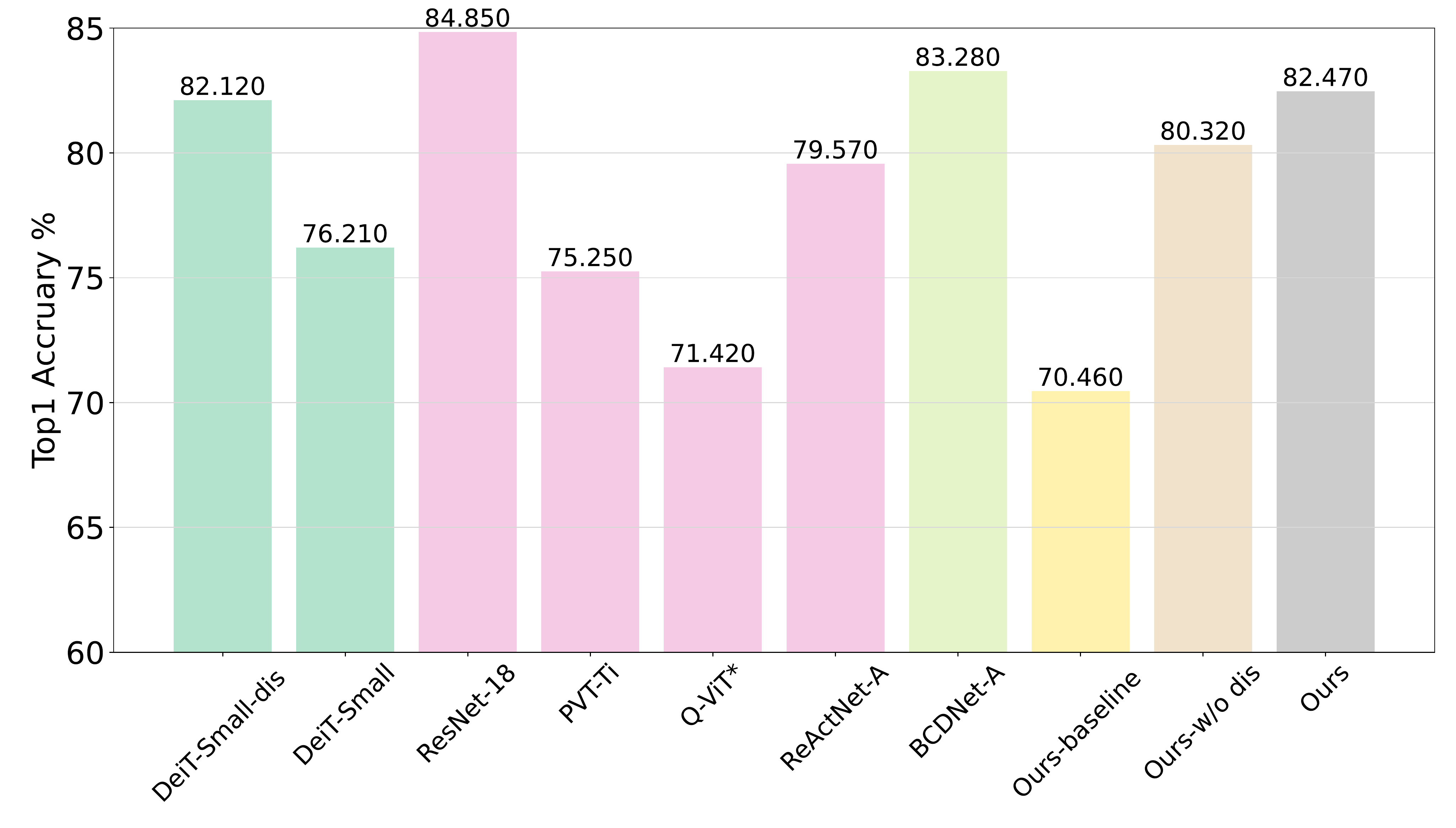}
\caption{Comparison results for classification on Chaoyang dataset.}
\label{figchaoyang}
\end{figure}

Compared with the Oxford-Flowers102 dataset, the ratio between the train set and the test set of the Chaoyang dataset is relatively reasonable. Moreover, the class number of Chaoyang is small. Therefore, DeiT-small without distillation shows quite good performance. Almost 40\% of Chaoyang dataset's training set has noisy labels and its test set is completely correct. This setting can help to evaluate the robustness of the model to label noise. As shown in Tab.~\ref{tab3} and Fig.~\ref{figchaoyang}, compared with CNN, the performance of the transformer based model is still weak. Compared with the baseline, 1 bit-Q-ViT*~\citep{li2022q} and ReActNet-A~\citep{liu2020reactnet}, our proposed method obtains outstanding performance and is close to the full-precision model with the same number of parameters. The number of parameters of BCDNet-A~\citep{xing2022towards} is almost twice that of our method, which provides the foundation for its excellent performance. This experiment shows that our binarization model not only has the ability to alleviate the overfitting problem but also is robust to label noise.
\begin{table}[htbp]
    	\caption{Comparison results for classification on Chaoyang dataset.}
		\label{tab3}
		\centering
		\begin{tabular}{c|c|c|c|c|c}  
			\hline
			\hline
			Methods                                  &Disti.    &Trans.& W-A   & Params number   &  Top-1(\%)          \\
			\hline
                \hline
			DeiT-Small~\citep{touvron2021training}                     &\ding{52}    &\ding{52}& 32-32   & 21.7M    & 82.12   \\
			\hline
   		    PVT-Ti~\citep{wang2021pyramid}                             &\ding{55}    &\ding{52}& 32-32  & 12.7M     & 75.25   \\
			\hline
			DeiT-Small~\citep{touvron2021training}                      &\ding{55}    &\ding{52}& 32-32  & 21.7M      & 76.21   \\
			\hline
			ResNet-18~\citep{he2016deep}                               &\ding{55}    &\ding{55} &32-32   & 11.3M     & \textbf{84.85}    \\
            \hline
            Q-ViT*~\citep{li2022q}                                     &\ding{52}    &\ding{52} & 1-1     & 21.7M    & 71.42    \\
            \hline
             ReActNet-A~\citep{liu2020reactnet}                                    &\ding{52}    &\ding{55} & 1-1     &28.3M   & 79.57  \\
            \hline
            BCDNet-A~\citep{xing2022towards}                                     &\ding{52}    &\ding{55} & 1-1    &44.1M    & 83.28   \\
            \hline
			Ours-baseline                                             &\ding{55}    &\ding{52} & 1-1    & 21.7M     & 70.46   \\
            \hline
			Ours-w/o disti                                            &\ding{55}    &\ding{52} & 1-1    & 24.5M     & 80.32   \\
            \hline
			Ours                                                      &\ding{52}    &\ding{52} & 1-1    & 24.5M     & \textbf{82.47}   \\
			\hline
			\hline
		\end{tabular}
	\end{table}
\subsection{Ablation Study}
\begin{figure}[htbp]
\setlength{\abovecaptionskip}{5pt}
\setlength{\belowcaptionskip}{0pt}
\centering
\includegraphics[width=5.0in]{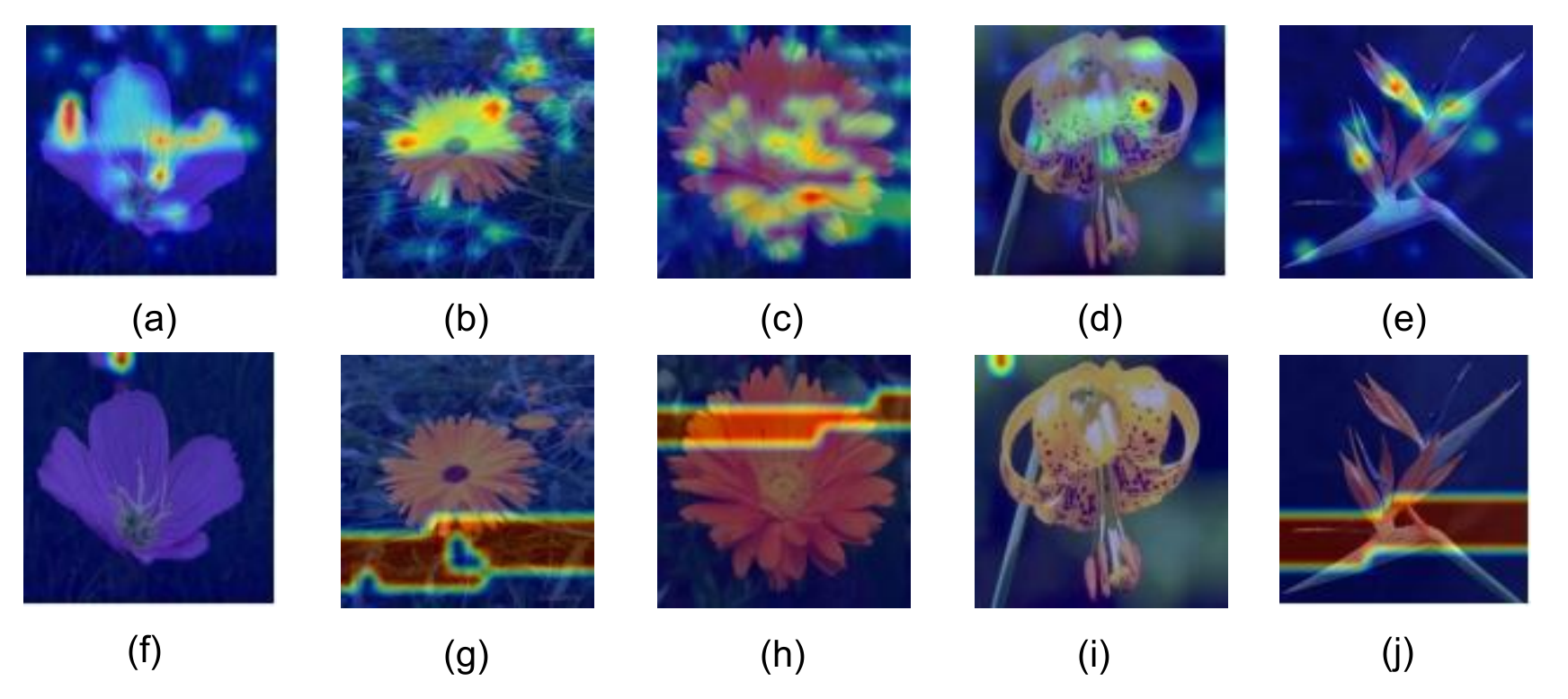}
\caption{The visualization of attention in the last Attention block between the class token and all image tokens w/wo using the GSB module. (a)-(e) represents the visualization of attention when using GSB. (f)-(j) represents the visualization of attention without GSB.}
\label{figattdb}
\end{figure}

\noindent \textbf{The effectiveness of each module on the CIFAR-100 dataset.} As shown in Tab.~\ref{tab4},  we first binarized Deit-Small with our baseline, resulting in a decrease in performance of 19.50\%. Then, we replace the binarization scheme of each module with GSB in turn, which results in an improvement of 15.81\%. Finally, we utilize the distillation method, which results in a performance improvement of 4.23\%. All modules have a positive effect on model performance. Modules that \textbf{GSB-Att} and \textbf{GSB-V} have a large effect on improving the performance of the binarization ViT model. Fig.~\ref{figattdb} represents the visualization of attention in the last Attention block between the class token and all image tokens w/wo using the GSB module, respectively. ALL images belong to the Oxford-flower102 dataset. The visualization code used is consistent with the method described in the ViT paper. From the Fig.~\ref{figattdb}, it can be observed that without GSB, there is a significant error in the model's attention mechanism. However, with GSB, the error in binarized attention decreases. The results of this comparative visualization validate the effectiveness of GSB scheme.
\begin{table}[htbp]
    	\caption{Ablation studies when training from scratch on CIFAR-100.}
		\label{tab4}
		\centering

		\begin{tabular}{c|c|cccc}  
			\hline
			\hline
			Methods                & W-A                    &GSB-att   &GSB-V      &$L_{dis}$     &  Top-1(\%)          \\
			\hline
                \hline
            DeiT~\citep{touvron2021training}   &32-32                   & ~         & ~        &\ding{52}              &70.56\\
            DeiT~\citep{touvron2021training}   &32-32                   & ~         & ~        &             &68.52  \\
			\hline
			\multirow{5}{*}{GSB}                    &\multirow{5}{*}{1-1}    &           &          &                       &51.06   \\
			
   		    ~                                       &~                       &\ding{52}  &         &                        &60.62       \\
			
			~                                       &~                       &\ding{52}  &\ding{52}&                        &66.87       \\

			~                                       &~                       &\ding{52}  &\ding{52}   &\ding{52}   &71.10       \\

			\hline
			\hline
		\end{tabular}
	\end{table}
 
\hspace*{\fill}

\noindent \textbf{The effectiveness of two binarization methods for different activation distribution.}  The distribution of the values of activations in Vison Transformer can be divided into two categories (\textit{type-A} and \textit{type-B}). Applying an inappropriate binarization operation will change the distribution of activations, eg. applying the $sign$ function to handle the activations of \textit{type-A}.
The distribution mismatch~\citep{liu2022bit} caused by the change in the distribution of activations before and after binarization between binarized activations and the corresponding full-precision activations will bring adverse effects, eg. The degradation of the performance of the model.


Inspired by BiT~\citep{liu2022bit}, we apply two binarization schemes with respect to different distributions of activations to solve the problem stated above. We explore the effectiveness of this scheme on vision tasks. In order to discard the influence of other factors, we only switch between the activation binarization schemes for our method. There are two settings: one is to apply only the $sign$ function to binarize all activations and the other is to use the $sign$ function and $round$ function to binarize according to the distribution type of the full-precision activation.  Meanwhile, we utilize the $sign$ function for the weight binarization (without applying distillation to ensure the fairness of the comparison). As shown in Tab.~\ref{tab6}, it is effective to use the $sign$ and $round$ functions to separately binarize the activations of different distributions. Furthermore, using the $sign$ function to complete the binarization process of the attention matrix (\textit{type-A}) will get an all-one matrix and destroy the function of the attention module.
\begin{table}[htbp]
    	\caption{Ablation studies about different settings for activation binarization on CIFAR-100. The weights of the model are binarized by the $sign$ function.}
		\label{tab6}

		\centering
		\begin{tabular}{c|c|c|c}  
			\hline
			\hline
			\multirow{2}{*}{Framework}     &\multicolumn{2}{c|}{Binarization Method}                            &\multirow{2}{*}{Top-1(\%).}       \\
            \cline{2-3}
            ~&\textit{type-A}&\textit{type-B}&~\\
			\hline
                \hline
			Ours-w/o disti        &    $sign$    & $sign$     &40.38     \\
			\hline
   		Ours-w/o disti     &   $round+clip$     &  $sign$               &66.87     \\
			\hline
			\hline
		\end{tabular}
	\end{table}
\hspace*{\fill} \\

\noindent \textbf{Ablation studies on the parameter} $k$. In this subsection, we focus on the influence of the number of additional components $k$ in the GSB binarization mechanism on the experimental results. We removed the distillation module during algorithm training. All models with different $k_{a}$ and $k_{v}$ for \textbf{GSB-Att} and \textbf{GSB-V}, respectively, are trained for 300 epochs with one stage in the Oxford-Flowers102 dataset from scratch. The result is shown in Tab.~\ref{tabK}.  From the results, we can see that the performance is generally positively correlated with the size of $k$. However, when $k$ is larger than 2, the performance improvement is very limited, and the computational complexity increases steadily. Considering the balance of these two aspects, we choose $k_{a}=2$, $k_{v}=2$ as the setting of the GSB ViT model.\\
\begin{table}[htbp]
\setlength{\abovecaptionskip}{2pt}
\caption{Ablation studies on the setting of the parameter $k$ for \textbf{GSB-Att} and \textbf{GSB-V}.}
\label{tabK}
\centering
\begin{tabular}{l|l|l|l|l|l}
\hline
\hline
\diagbox{$k_{v}$}{Top-1(\%)}{$k_{a}$}   & $k_{a}=0$ & $k_{a}=1$ & $k_{a}=2$ & $k_{a}=3$ & $k_{a}=4$ \\
\hline
\hline
$k_{v}=0$ & 35.48     & 42.34&43.62&43.76& 43.88     \\
\hline
$k_{v}=1$ & 38.27     & 42.86     & 44.05     &  43.95    &  44.36    \\
\hline
$k_{v}=2$ &40.42      &43.26      &\textbf{44.89}&44.75&44.96     \\
\hline
$k_{v}=3$ & 41.25     & 43.71     &  44.91    & 44.95     &  45.01    \\
\hline
$k_{v}=4$ & 41.33     &  43.75 &45.07& 45.22  &  45.63    \\
\hline
\hline
\end{tabular}
\end{table}

\noindent \textbf{Ablation studies on the offset $\boldsymbol{\Phi}$ of real-valued attention matrix.} In this subsection, we conducted experiments to study the impact of learnable $\boldsymbol{\Phi}$ with different settings on model performance. The CIFAR-100 dataset is utilized to train the GSB ViT model with different offsets in this ablation study. The result is shown in Tab.~\ref{taboffset}. Four different settings of $\boldsymbol{\Phi}$ are tested in this ablation study. The first is the same as our baseline by setting the offset to 0. The second is the mean value of $\mathbf{A}_{re}$. The remaining settings are learnable offset with different tensor sizes. Compared with the corresponding component in the full-precision network, the dot product of the binarized $\mathbf{Q}$ and binarized $\mathbf{K}$ has a non-negligible error. To address this issue, we introduce learnable offsets to fix this error term. From the result, compared with the setting of $\boldsymbol{\Phi}=0$, we can find out that the learnable offset has a positive effect on the performance of the model. With the increase of the number of parameters of offset, the convergence speed and performance upper limit of the model have been clearly improved. Because the offset only performs addition or subtraction operations, its parameter size does not affect the computational complexity of the forward inference of the model although it affects the storage space requirements of the model to an acceptable extent.
 \begin{table}[htbp]
    	\caption{Ablation studies about different settings for offset $\boldsymbol{\Phi}$ of GSB-ViT model.}
		\label{taboffset}
		\centering
		\begin{tabular}{c|c|c|c}  
			\hline
			\hline
			Type  & Offset          &  Epochs       &          Top-1(\%).       \\
			\hline
            \hline
			\multirow{2}{*}{Unlearnable}  &  $\boldsymbol{\Phi}=0$     &    900        &67.22     \\
            \cline{2-4}
            ~  &  $\boldsymbol{\Phi}=mean\left(A_{re}\right)$    &    900        &56.47\\
			\hline
   		\multirow{2}{*}{Learnable}   & $\boldsymbol{\Phi}\in \mathbb{R}^{H\times N}$   &   900                   &68.92     \\
            \cline{2-4}
            ~  &  $\boldsymbol{\Phi}\in \mathbb{R}^{H\times N\times N}$    &    300        &71.10     \\
			\hline
			\hline
		\end{tabular}
	\end{table}

\begin{figure}[htbp]
\setlength{\abovecaptionskip}{5pt}
\setlength{\belowcaptionskip}{0pt}
\centering
\includegraphics[width=5.0in]{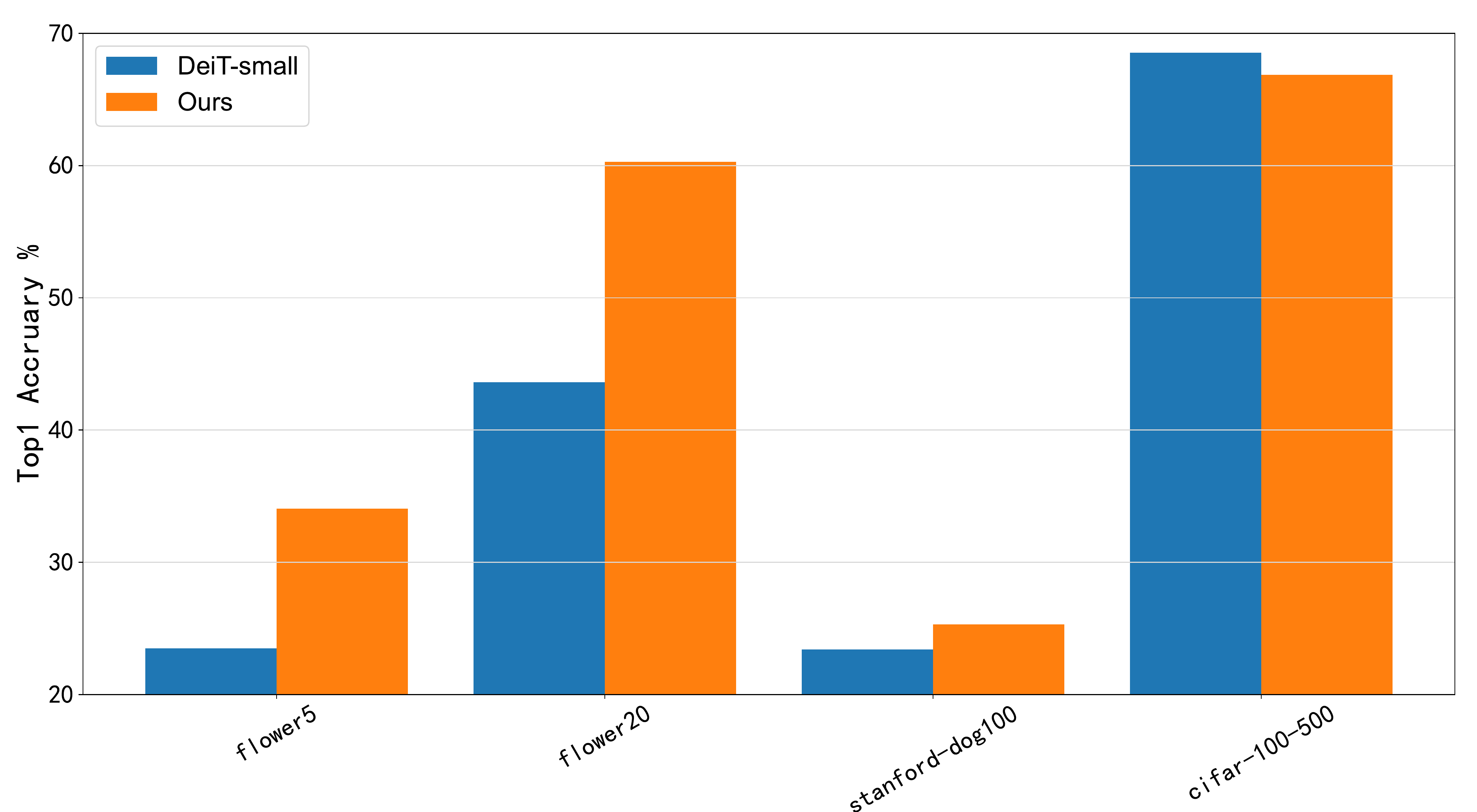}
\caption{Comparison of performance between GSB method and full-precision Deit-small under different training data sizes.}
\label{figduibi}
\end{figure}

\noindent \textbf{Ablation studies on the definition of $\mathbf{A}_{bin}$.}  Compared with $\mathbf{V}_{bin}^{GSB}$, the equation of $\mathbf{A}_{bin}^{GSB}$ is more complex. The reason is that the definition of the components in $\mathbf{A}_{bin}^{GSB}$ is complicated. From the Eq.~\ref{eq5}, Eq.~\ref{eq5-1}, and Eq.~\ref{eq5-2}, we can find out that the definition of $\mathbf{A}_{bin}$ is different from $\mathbf{M}_{i}$. In order to verify the effectiveness of our special definition for $\mathbf{A}_{bin}$, we conduct ablation experiments on $\mathbf{A}_{bin}$ in this subsection. The result is listed in Tab.~\ref{tababin}, which shows that our setting is more useful and has a positive effect on the performance of the model.\\
 \begin{table}[htbp]
    	\caption{Ablation studies about the model with different defination of $\mathbf{A}_{bin}$ trained on CIFAR-100 dataset.}
		\label{tababin}
		\centering
        \setlength{\tabcolsep}{9mm}{
		\begin{tabular}{c|c}  
			\hline
			\hline
			The defination of $\mathbf{A}_{bin}$          &Top-1(\%).       \\

			\hline
                \hline
			Same as $\mathbf{M}_{i}$            &59.28     \\
			\hline
   		Ours            &71.10     \\
			\hline
			\hline
		\end{tabular}}
	\end{table}\\
\noindent \textbf{Ablation studies on the size of train dataset.} In this subsection, we test the performance of our method and the corresponding full-precision model on different amounts of training data. The datasets include subsets of Oxford Flower-120, Stanford Dogs~\cite{dataset2011novel}, and CIFAR-100. In the datasets flower5, flower20, stanford-dog100, and cifar-100-500, the training data quantities per category are 5, 20, 100, and 500, respectively. These datasets have similar numbers of classes but varying levels of classification difficulty. To ensure a fair comparison, no distillation techniques were used in the training processes of both models, and their hyperparameters were kept the same. The results are shown in Fig.~\ref{figduibi}.
From the Fig.~\ref{figduibi}, we observed that when the number of training samples is less than 100, the full-precision Deit-small model performs poorly due to overfitting issues. However, our binarization algorithm mitigates the impact of overfitting. When the number of training samples is more than 100, the full-precision model slightly outperforms our binarized model due to binarization errors. The experimental results indicate that our algorithm can alleviate the overfitting effect in scenarios with limited training data and deliver relatively preferred performance.
\subsection{Computational Complexity Analysis}
Following the TNT~\citep{han2021transformer}, we can calculate the FLOPs of a full-precision transformer block by the Eq.~\ref{eqflops}. \\
\begin{equation}
\begin{aligned}
\label{eqflops}
F_{MHA}&=2nd\times \left( 2d+n \right), \\
F_{MLP}&=2ndrd,
\end{aligned}
\end{equation}
where $F_{MHA}$ and $F_{MLP}$ indicate the FLOPs of the multi-head attention module and the MLP in a transformer block, respectively. $n$ and $d$ are the number of image patches and the number of dimensions, respectively. $r$ is the dimension expansion ratio of the hidden layer in MLP. \\
\indent Since the activation binarization operation cannot be avoided, there are still some FLOPs in the forward propagation process of the binarized model. For the fairness of the comparison, we follow the scheme proposed in~\citep{liu2020reactnet} to calculate the number of operations (OPs) of the binary network and the corresponding full-precision one. The definition of OPs is $\mathbf{OPs=\frac{BOPs}{64}+FLOPs}$. BOPs and FLOPs mean binary operations and floating point operations, respectively. We provide an analysis of the requirement of computing resources for each transformer block in Tab.~\ref{tab5}. Compared with the DeiT-Small model~\citep{touvron2021training}, the proposed baseline model and the GSB ViT increase the operation efficiency by about 37.2 times and 30 times in terms of OPs, respectively.
\begin{table}[htbp]
    	\caption{Comparison results for the computational complexity of one transformer block. All OPs is equal to $\left(\mathbf{\frac{BOPs}{64}+FLOPs}\right)$.}
		\label{tab5}
		\centering
        \scalebox{0.95}{
		\begin{tabular}{c|c|c|c|c|c}  
			\hline
			\hline
			\multirow{3}{*}{Methods}                &\multicolumn{2}{c|}{Attention}   &\multicolumn{2}{c|}{MLP}     &  \multirow{3}{*}{All OPs }         \\
            \cline{2-5}
                                                                                                    &    BOPs            & FLOPs  &    BOPs            & FLOPs        \\
             ~       &    $\left(\times 10^6\right)$            & $\left(\times 10^6\right)$  &    $\left(\times 10^6\right)$            & $\left(\times 10^6\right)$   & $\left(\times 10^6\right)$    \\
			\hline
                \hline
			DeiT-Small~\citep{touvron2021training}                        &-& 147          & -     & 233       &380  \\
			\hline
			Ours-baseline                                                 &147 & 2.76     & 233  &1.52       & 10.21  \\
            \hline
			Ours                                                         &267 & 3.93     & 233  &1.14      & 12.88  \\
			\hline
			\hline
		\end{tabular}}
	\end{table}

\section{Discussion and Conclusion}\label{section6}
\noindent \textbf{Discussion}: The findings of this study have to be seen in light of some limitations. Firstly, our GSB method on attention matrix does introduce additional computational complexity. However, the increase in computation is justified by the improved representation capacity of the binarized attention matrix. Currently, there is no perfect solution to reconcile this trade-off in the existing research. Therefore, in future work, we will explore attention structures that are suitable for binarized models to mitigate this impact. Secondly, we have not conducted actual evaluations on hardware with limited computational resources such as microcontrollers or FPGAs. Although this issue does not affect the algorithm design and theoretical analysis, the applicability of the algorithm is still relatively important. Therefore, specific hardware deployment will be completed in future work to ensure the practical feasibility of the approach.
\noindent \textbf{Conclusion}: In this paper, we propose a novel binarized ViT method, called Group Superposition Binarization, to deal with the overfitting problems of common ViT models when training with limited samples, and simultaneously improve the efficiency of ViT models. Compared with the full-precision networks, the proposed method reduces the computational complexity by compressing the weight and activation of the model to the 1-bit level. We deduce the gradient of GSB to reduce the performance degradation caused by the gradient mismatch from binarizing the model.
Analytically, the proposed GSB binarization can limit
the parameter’s search space during parameter updates. Essentially, the binarization process can actually play an implicit regularization role, and thus it is helpful in dealing with the problem of
overfitting in the case of insufficient training data.
Experimental results on three datasets with limited training samples have demonstrated the effectiveness of the proposed method with  insufficient data and even label noise.


\section*{Acknowledgements}
This research did not receive any specific grant from funding agencies in the public, commercial, or not-for-profit sectors.


\bibliographystyle{elsarticle-harv}
\bibliography{egbib}







\end{document}